\RequirePackage{fix-cm}

\documentclass[twocolumn]{svjour3}          
\usepackage{tcolorbox}
\usepackage{tikz}
\usepackage{tcolorbox}
\usepackage{xcolor}
\usepackage[export]{adjustbox}
\usepackage{graphicx,color}
\graphicspath{ {./ImagesJPG/} }
\usepackage[latin5]{inputenc}
\usepackage[T1]{fontenc}

\usepackage{graphicx}
\usepackage{subfigure}
\usepackage{multirow}
\usepackage{array}
\begin{document}

\title{Hybrid Light Field Imaging for Improved Spatial Resolution and Depth Range \thanks{This work is supported by TUBITAK Grant 114E095.}}

\author{M. Zeshan Alam     \and
        Bahadir K. Gunturk
}

\institute{The authors are with \at
              Istanbul Medipol University, Istanbul, Turkey \\
              \email{mzalam@st.medipol.edu.tr}
              \email{bkgunturk@medipol.edu.tr}
}

\date{Received: date / Accepted: date}

\maketitle

\begin{abstract}
Light field imaging involves capturing both angular and spatial distribution of light; it enables new capabilities, such as post-capture digital refocusing, camera aperture adjustment, perspective shift, and depth estimation. Micro-lens array (MLA) based light field cameras provide a cost-effective approach to light field imaging. There are two main limitations of MLA-based light field cameras: low spatial resolution and narrow baseline. While low spatial resolution limits the general purpose use and applicability of light field cameras, narrow baseline limits the depth estimation range and accuracy. In this paper, we present a hybrid stereo imaging system that includes a light field camera and a regular camera. The hybrid system addresses both spatial resolution and narrow baseline issues of the MLA-based light field cameras while preserving light field imaging capabilities.
\end{abstract}

\section{Introduction}

A light field can be defined as the collection of all light rays in 3D space \cite{gershun,levoy}. One of the earliest implementations of a light field camera was presented \cite{lippmann}, where a micro-lens array is placed in front of a film to capture incident light amount from different directions. While a light field, in general, can be parameterized in terms of 3D coordinates of ray positions, 2D ray directions, and physical properties of light  (such as wavelength and polarization), the independent parameters can be reduced to a four-dimensional space assuming there is no energy loss during light propagation and when only the intensity of light is considered; such a four-dimensional representation of light field is used in many practical applications \cite{Adelson,levoy,Gortler}. Unlike regular cameras, light field cameras capture the directional light information, which enables new capabilities, including post-capture adjustment of camera parameters (such as focal length and aperture size), post-capture change of camera viewpoint, and depth estimation. As a result, light field imaging is getting increasingly used in a variety of application areas, including digital photography, microscopy, robotics, and machine vision.

Light field imaging systems can be implemented in various ways, including camera arrays \cite{Wilburn,levoy,yang}, micro-lens arrays \cite{Ng,Lumsdaine}, coded masks \cite{Veeraraghavan}, objective lens arrays \cite{Georgiev}, and gantry-based camera systems \cite{Unger}. Among these different implementations, micro-lens array (MLA) based light field cameras offer a cost-effective approach; and it is widely adopted in academic research as well as in commercial light field cameras \cite{lytro,raytrix}.

MLA-based light field cameras have two limiting issues. The first one is low spatial resolution. Because the image sensor is shared to capture both spatial and angular information, MLA-based light field cameras suffer from a fundamental resolution trade-off between spatial and angular resolution. For example, the first-generation Lytro camera has a sensor of around 11 megapixels, producing 11x11 angular resolution and less than 0.15 megapixel spatial resolution \cite{dansereau}. The second-generation Lytro camera has a sensor of 40 megapixels; however, this large resolution capacity translates to only four megapixel spatial resolution (with the manufacturer's decoding software) due to the angular-spatial resolution trade-off.

The second issue associated with MLA-based light field cameras is narrow baseline. The distance between sub-aperture images decoded from a light field capture is very small, significantly limiting the depth estimation range and accuracy. For instance, the maximum baseline (between the leftmost and rightmost sub-aperture images) of a first-generation Lytro camera is less than a centimeter, which typically results in sub-pixel feature disparities. There are methods in the literature specifically designed to estimate disparities and depth maps for MLA-based light field cameras \cite{Yu,Jeon,Tao}.

\begin{figure*}
  \includegraphics[width=6in]{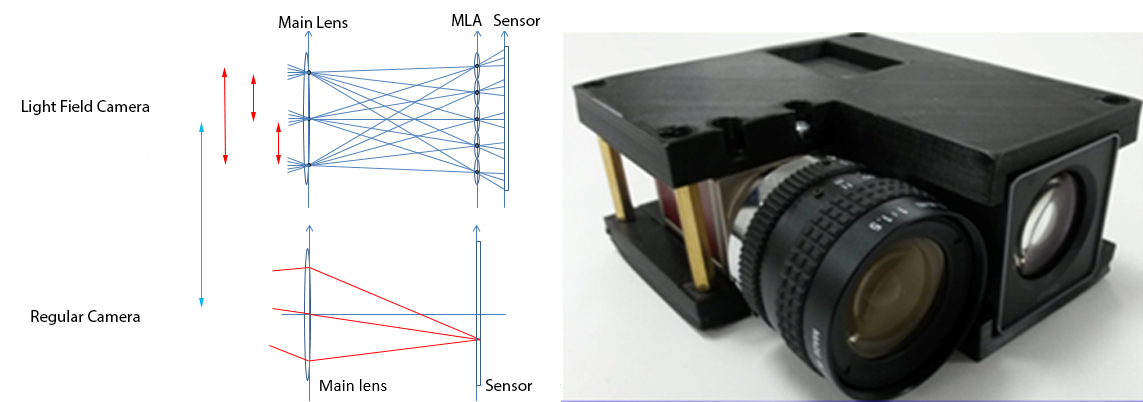}
  \centering
  \caption{Hybrid imaging system including a regular and a light field camera. The maximum baseline of the light field camera is limited by the camera main lens aperture, and is much less (about an order of magnitude) than the baseline (about 4cm) between the light field and the regular camera.}
  \label{fig:fig1}
\end{figure*}

To address both resolution and baseline issues, we propose a hybrid stereo imaging system that consists of a light field camera and a regular camera.\footnote{A preliminary version of this work was presented as a conference paper \cite{Alam}. In this paper, we provide additional experiments, detailed algorithm explanation and analysis.} The proposed imaging system is shown in Figure \ref{fig:fig1}; it has two main advantages over a single light field camera: First, high spatial resolution image captured by the regular camera is fused with low spatial resolution sub-aperture images of the light field camera to enhance the spatial resolution of each sub-aperture image; that is, a high spatial resolution light field is obtained while preserving the angular resolution. Second, the distance between the light field camera and the regular camera produces a larger baseline compared to the maximum baseline of the light field camera; as a result, the hybrid system has better depth estimation range and accuracy.

Hybrid light field imaging systems have been presented previously \cite{Boominathan,wu,wang}. Unlike these previous work, where spatial resolution enhancement is the only goal, we achieve wider baseline through using a calibrated system. With wide baseline, both range and accuracy of depth estimation are improved in addition to spatial resolution enhancement. Fixing cameras as a stereo system also enables offline calibration and rectification, reducing the computational cost.

In Section 2, related work addressing spatial resolution and narrow baseline issues of MLA-based light field cameras is provided. The proposed hybrid imaging system with resolution enhancement algorithm is explained in Section 3. Experimental results on resolution enhancement and depth estimation are presented in Section 4. Concluding remarks are given in Section 5.

\section{Related Work}

{\it On low spatial resolution:} There are various methods proposed to address the low spatial resolution issue in MLA-based light field cameras. One main approach is to apply super-resolution image restoration to light field sub-aperture images. Super-resolution in a Bayesian framework is commonly used, for example, in \cite{Bishop} with Lambertian and textural priors, in \cite{mitra} with a Gaussian mixture model, and in \cite{Wanner} with a variational formulation. Learning-based methods are adopted as well, including dictionary-based learning  \cite{Cho} and deep convolutional neural networks \cite{Yoon,Kalantari}. In addition to spatial domain super-resolution restoration, Fourier-domain techniques \cite{Perez,Shi} and wave optics based 3D deconvolution methods \cite{Shroff,Broxton,Sevilla,Junker} have also been utilized.

Alternative to the standard MLA-based light field camera design \cite{Ng}, where the MLA is placed at the image plane of the main lens and the sensor is placed at the focal length of the lenslets, there is another design approach where the MLA is placed to relay image from the intermediate image plane of the main lens to the sensor \cite{Lumsdaine}. This design is known as ``focused plenoptic camera.'' As in the case of the standard light field camera approach, super-resolution restoration for focused plenoptic cameras is also possible \cite{Georgiev2009}.

All single-sensor light field imaging systems are fundamentally limited by the spatial-angular resolution trade-off, and the above-mentioned restoration methods have performance limitations in addition to the computational costs. Another approach for improving spatial resolution is to use a hybrid two camera system, including a light field camera and a high-resolution camera, and merge the images to improve spatial resolution \cite{Boominathan,wu,wang}. Dictionary-learning based techniques are adopted \cite{Boominathan,wu} in this problem as well: High-resolution image patches from the regular camera are extracted and stored as a high-resolution patch dictionary. These high-resolution patches are downsampled; and from the downsampled patches, low-resolution features are extracted to form a low-resolution patch dictionary. During super-resolution reconstruction, a low-resolution image patch is enhanced through determining (based on feature matching) and using the corresponding high-resolution patches in the dictionary. In \cite{wang}, high-resolution image is decomposed with complex steerable pyramid filters; the depth map from the light field is upsampled using joint bilateral upsampling; perspective shift amounts are estimated from the upsampled depth map, and these shift amounts are used to modify the phase of the decomposed high-resolution image; with the modified phases, pyramid reconstruction is applied to obtain high-resolution light field.

{\it On narrow baseline:} One of the most important features of light field cameras is the ability to estimate depth. However, it is known that depth accuracy and range is limited in MLA-based light field cameras due to narrow baseline. The relation between baseline and depth estimation accuracy in a stereo system has been studied in \cite{Gallup}. In a stereo system with focal length $f$ and baseline $b$, the depth $z$ of a point with disparity $d$ is obtained through triangulation as $z=fb/d$. With a disparity estimation error of $\epsilon_{d}$, the depth estimation error $\epsilon_{z}$ becomes \cite{Gallup}:

\begin{equation}
\epsilon_{z} = \frac{fb}{d} - \frac{fb}{d+\epsilon_{d}} = \frac{d^{2} \epsilon_{d}}{fb + d \epsilon_{d}} \approx \frac{z^{2}}{f b}\epsilon_{d},
\label{eq:5}
\end{equation}
which indicates that the depth estimation error is inversely proportional with the baseline and increases quadratically with depth. The disparity error $\epsilon_{d}$ is typically set to 1, and the depth estimation error $\epsilon_{z}$ as a function depth can be calculated. It is also possible to set an error bound on $\epsilon_{z}$ and derive the maximum depth range from the above equation.

For an MLA-based light field camera, the maximum baseline is less than the size of the main lens aperture, making depth estimation very challenging. There are methods specifically proposed for depth estimation in MLA-based light field cameras. For example, in \cite{Wanner2012}, the problem is formulated as a constrained labeling problem on epipolar plane images in a variational framework. In \cite{Yu}, ray geometry of 3D line segments is imposed as constraints on light field triangulation and stereo matching. In \cite{Tao}, defocus and shading cues are used to improve the disparity estimation accuracy.

\section{Hybrid Stereo Imaging}

The hybrid stereo imaging system consists of a regular camera and a light field camera as shown in Figure \ref{fig:fig1}. The system has two advantages over a single light field camera: (i) The high-resolution image produced by the regular camera is used to improve the spatial resolution of each sub-aperture image extracted from the light field camera. That is, we obtain a light field with enhanced spatial resolution. (ii) The large baseline between the regular camera and the light field camera results in a wider range and more accurate depth estimation capability, compared to a single light field camera.

\begin{figure}
\centering
\includegraphics[width=1.5in]{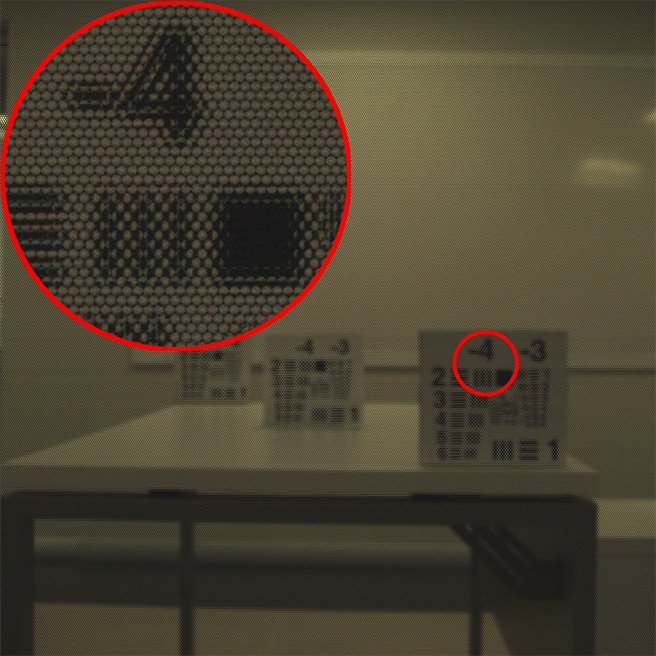}
\includegraphics[width=1.5in]{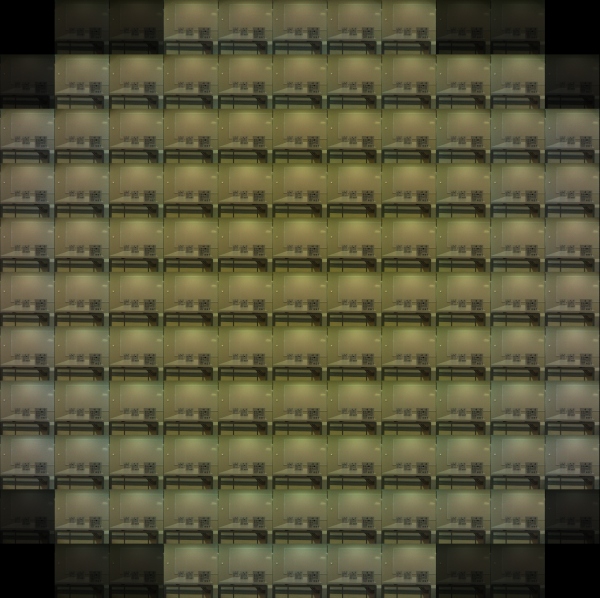}\\
(a) \hspace{1.3in} (b)
\caption{(a) Raw light field. (b) Decoded sub-aperture images.}
\label{fig:raw}
\end{figure}

\begin{figure}[h!]
\centering
\includegraphics[width=1.0in]{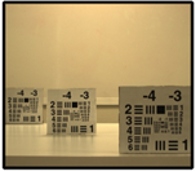}
\includegraphics[width=1.0in]{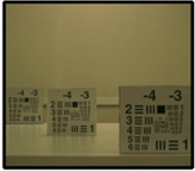}
\includegraphics[width=1.0in]{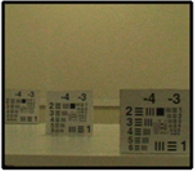}\\
(a) \hspace{0.8in} (b) \hspace{0.8in} (c)
\caption{(a) Regular camera image. (b) Regular camera image after photometric registration. (c) One of the bicubically resized Lytro sub-aperture image.}
\label{fig:photomet}
\end{figure}

\subsection{Prototype System and Initial Light Field Data Processing}

\begin{figure*}
\centering
\includegraphics[width=6in]{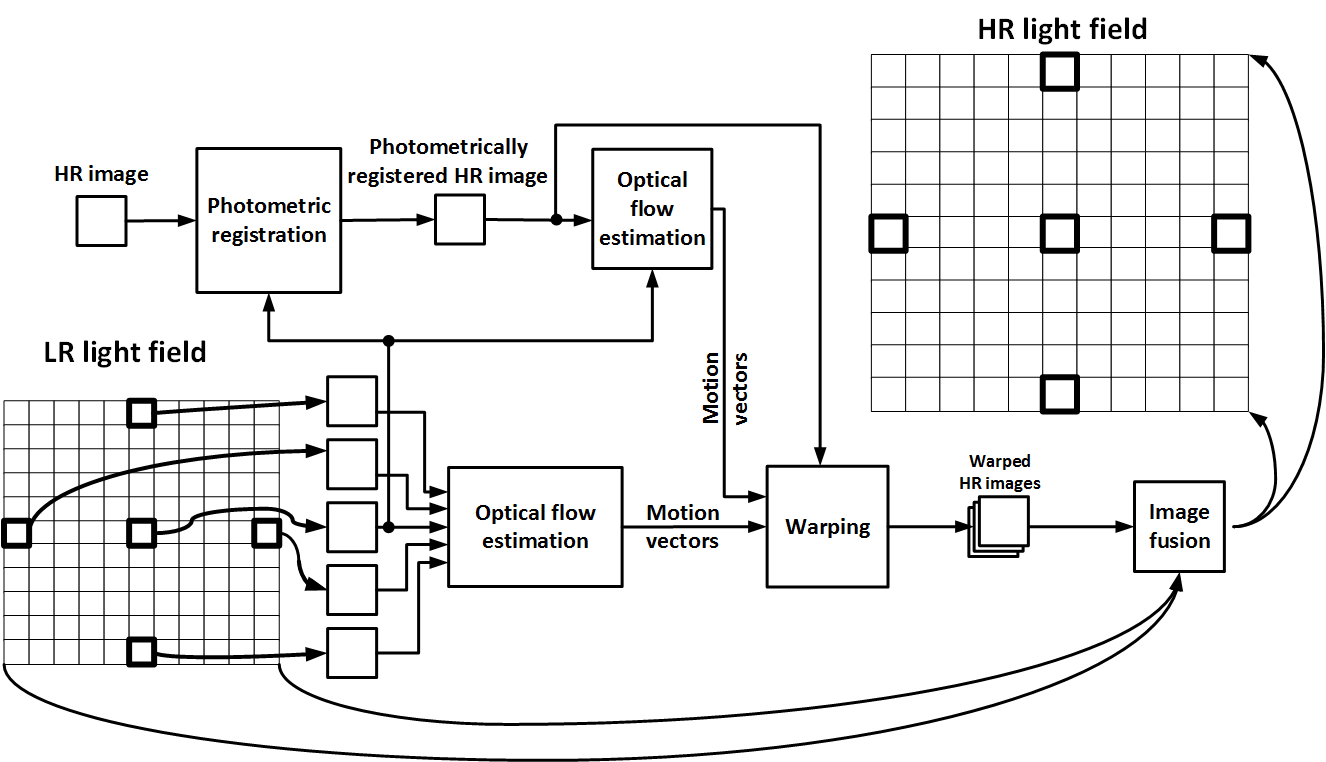}
\caption{Illustration of the resolution enhancement process.}
\label{fig:flowchart}
\end{figure*}

The prototype system includes a first-generation Lytro camera and a regular camera (AVT Mako G095C). The light field is decoded using \cite{dansereau} to obtain 11x11 sub-aperture images, each with size 380x380. The regular camera has a spatial resolution of 1200x780 pixels. The imaging system is first calibrated: The regular camera image and the light field middle sub-aperture image is calibrated (utilizing the Matlab Stereo Calibration Toolbox) to determine the overlapping regions between the images and rectify the regular camera image. The regular image is then photometrically mapped to the color space of the light field sub-aperture images using the histogram-based intensity matching function technique \cite{grossberg}.

A raw light field data and the extracted sub-aperture images are shown in Figure \ref{fig:raw}. In Figure \ref{fig:photomet}, the rectified regular camera image is shown along with a light field sub-aperture image.

\subsection{Improving Spatial Resolution}

An illustration of the resolution enhancement process is given in Figure \ref{fig:flowchart}. Each low-resolution (LR) light field sub-aperture image is bicubically interpolated to match the size of the high-resolution (HR) regular camera image. The optical flow between the HR image and the light field middle sub-aperture image and the optical flow between the light field middle sub-aperture and every other sub-aperture images are estimated. (We use the optical flow estimation algorithm presented in \cite{Liu} in our experiments.) Combining these optical flow estimates, motion vectors between the HR image and each of the light field sub-aperture images are obtained. The HR image is warped onto each light field sub-aperture image and fused to produce a high-resolution version of each sub-aperture image. As a result, a high-resolution light field is obtained.

Image fusion for resolution enhancement is a well-studied topic; the application areas include satellite imaging for pan-sharpening, digital camera pipelines for demosaicking, and computational photography for focus stacking \cite{Mitchell}. In our experiments, we tested two basic methods for image fusion: (i) a wavelet-based approach \cite{Zeeuw}, available in Matlab as function {\it wfuseimg}, which essentially replaces the detail subbands of low-resolution image with the detail subbands of the high-resolution image, and (ii) alpha blending, also available in Matlab as function {\it imfuse}, which simply takes the weighted average of input images.

\begin{figure}
   \centering
  \includegraphics[width=3.5in]{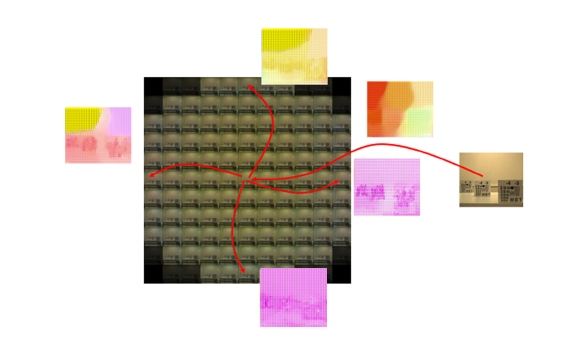}
  \caption{Speeding up the optical flow estimation process.}
  \label{fig:optical}
\end{figure}

\begin{figure}
  \centering
  \includegraphics[width=3in]{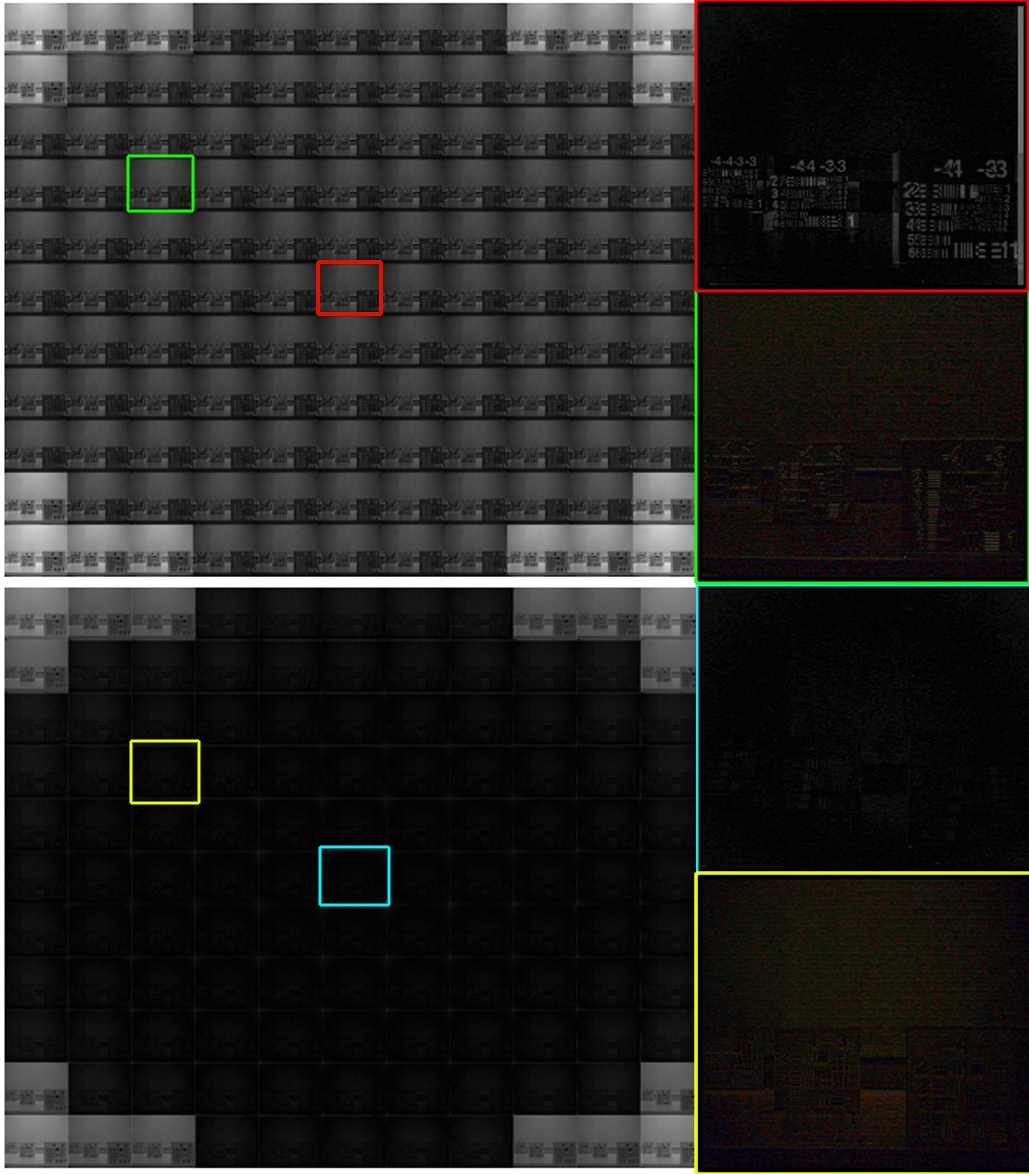}
  \caption{(Top) Residual between the regular camera image and light field sub-aperture images before warping. Two sub-aperture images are highlighted. (Bottom) Residual between the regular camera image and light field sub-aperture images after warping.}
  \label{fig:difference}
\end{figure}

We further increase the speed of registration process by using the fact that light field sub-aperture images are captured on a regular grid. Instead of estimating the optical flow between the middle sub-aperture image and each of the remaining sub-aperture images, we estimate the optical flow between the middle and the leftmost, rightmost, topmost, and bottommost sub-aperture images as shown in Figure \ref{fig:optical}. The estimated motion vectors are interpolated for the rest of the sub-aperture images according to their relative positions within the light field. As a result, we conduct four within-light-field-camera optical flow estimation (instead of 120) and one between-cameras optical flow estimation. In Figure \ref{fig:difference}, we show the difference between the regular camera image and light field sub-aperture images before and after registration. The optical flow within the light field is estimated as described above. The after-registration result shows that the registration process works well. (Note that the residuals for the sub-aperture images in the aperture corners are large because of the fact that the original sub-aperture images in the corners are too dark due to vignetting.)

\section{Experimental Results}

\begin{figure*}
\centering
\includegraphics[trim={0 0 0 2cm},clip,width=2.0in]{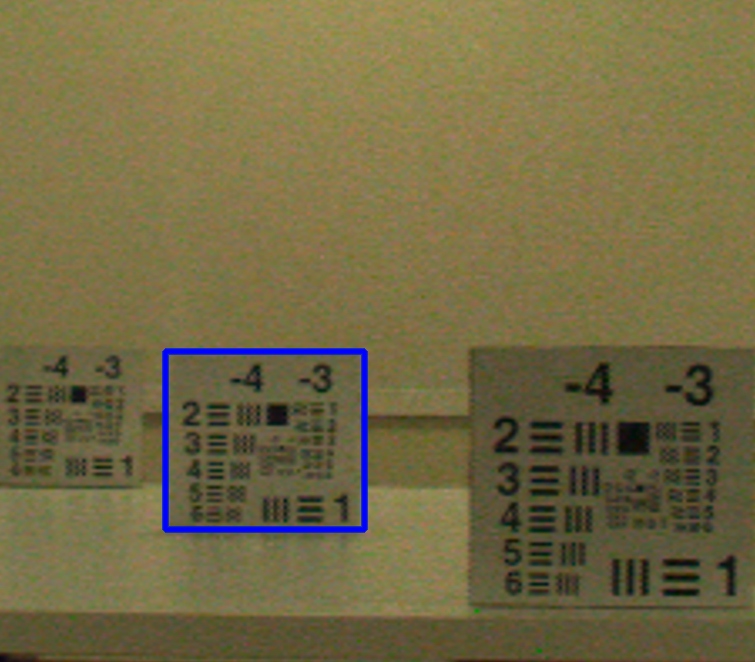}
\includegraphics[trim={0 0 0 2cm},clip,width=2.0in]{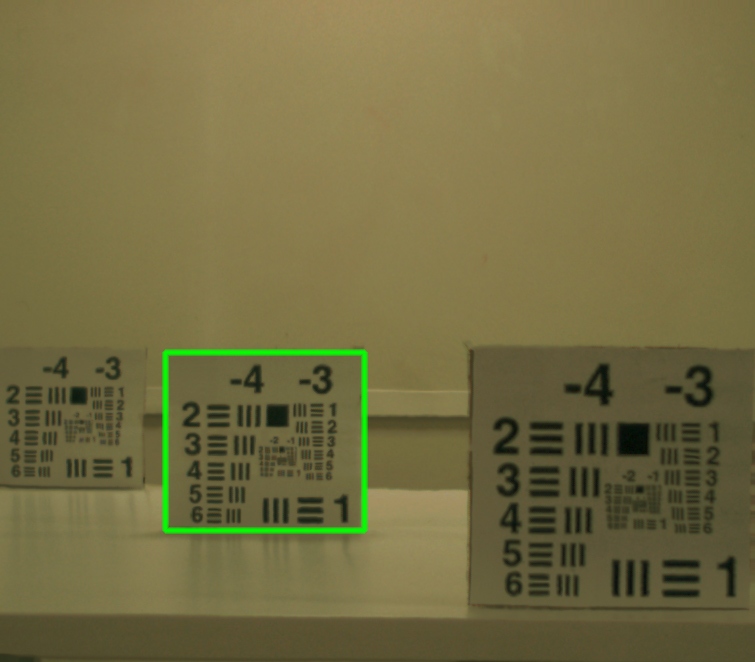}
\includegraphics[trim={0 0 0 2cm},clip,width=2.0in]{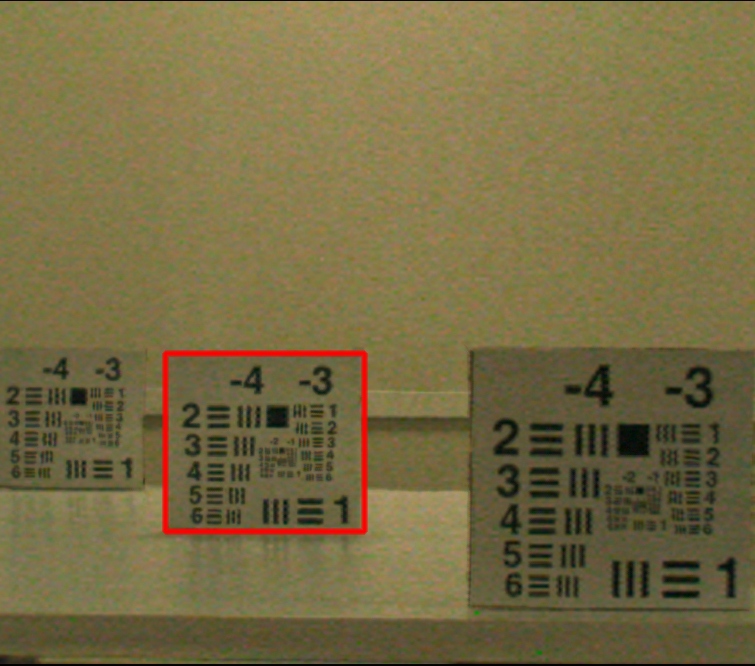}\\
\includegraphics[width=2.0in]{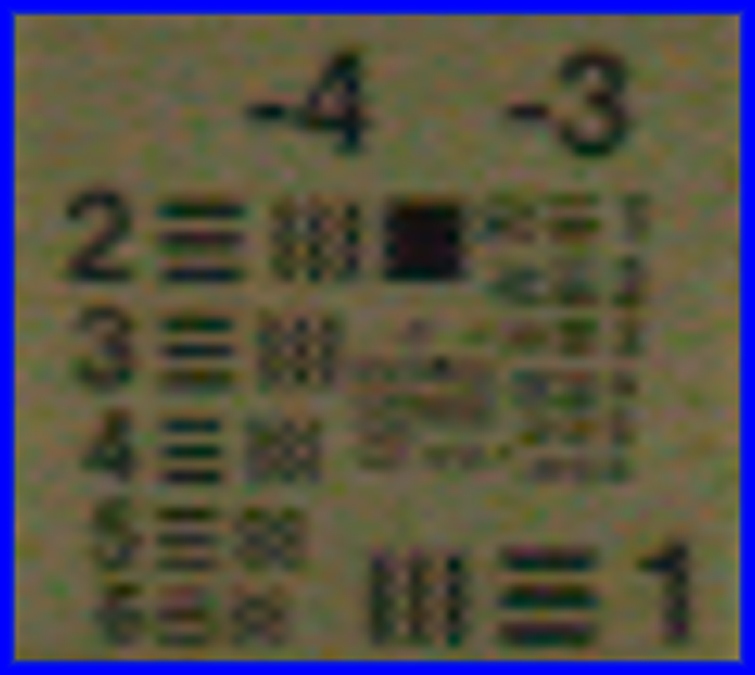}
\includegraphics[width=2.0in]{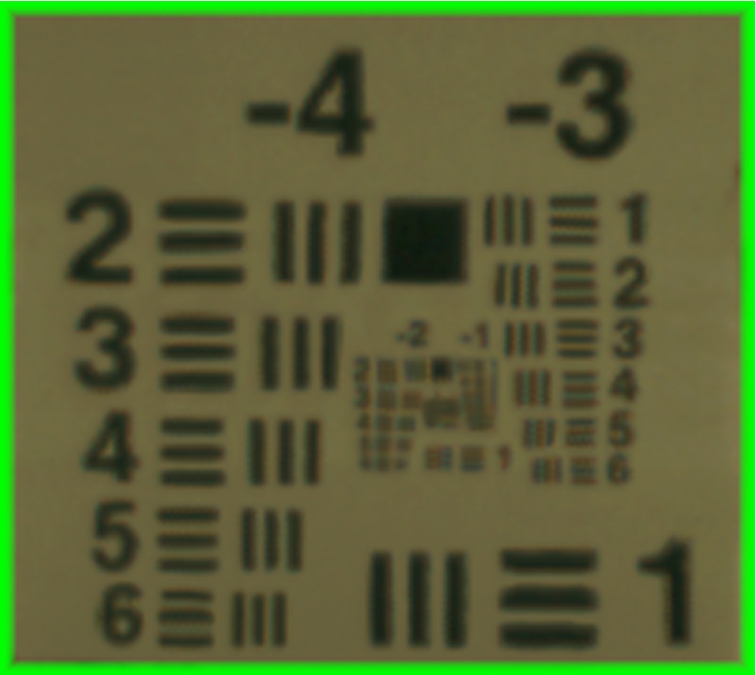}
\includegraphics[width=2.0in]{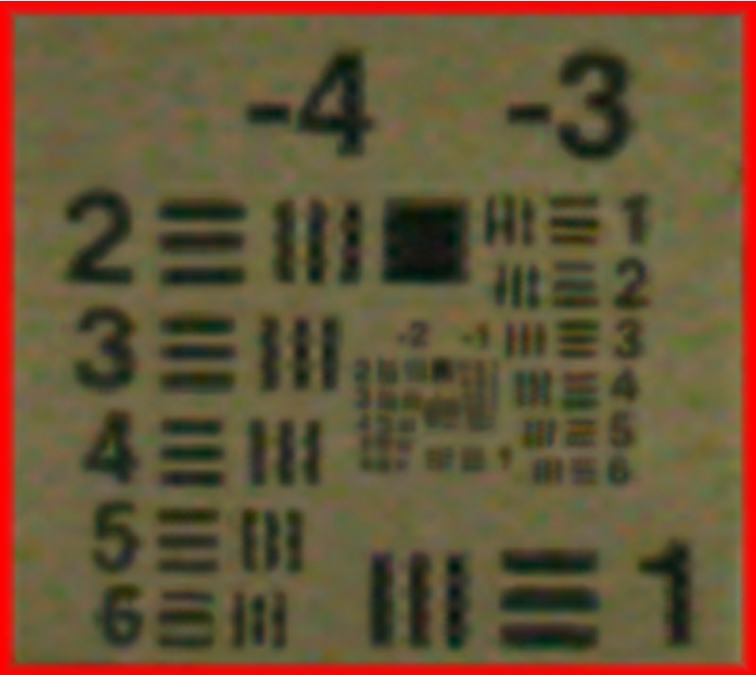}\\
(a) \hspace{1.8in} (b) \hspace{1.8in} (c)\\
\caption{Resolution enhancement of light field sub-aperture images. (a) One of the bicubically resized Lytro sub-aperture image. (b) Resolution-enhanced sub-aperture image using alpha blending. (c) Resolution-enhanced sub-aperture image using wavelet-based fusion.}
\label{fig:Result1}
\end{figure*}

In this section, we present our experimental results on resolution enhancement and depth estimation. All implementations are done with Matlab, running on an Intel i5 PC with 12GB RAM. For the resolution enhancement process, given in Figure \ref{fig:flowchart}, the processing time of an entire Lytro light field is about 70 seconds, in which the optical flow estimation per image pair is about 11 seconds. In Figure \ref{fig:Result1}, we compare a light field sub-aperture image with its resolution-enhanced version. Both the alpha blending and wavelet-based image fusion processes produce good results in terms of resolution enhancement. Alpha blending suppresses the low-spatial-frequency color noise better than the wavelet-based approach; this is expected because the wavelet-based approach preserves the low-frequency content of the light field images, which have more noise compared to the image obtained from the regular camera. Alpha blending, on the other hand, simply averages two images, reducing the overall noise in all parts of the final image. (In all our experiments, the weights of the HR image and light field sub-aperture images are 0.55 and 0.45, respectively, giving slightly more weight to the HR image in alpha blending.)

\begin{figure*}
\begin{subfigure}{}
\includegraphics [trim={0 0 0 0},width=0.24\textwidth]{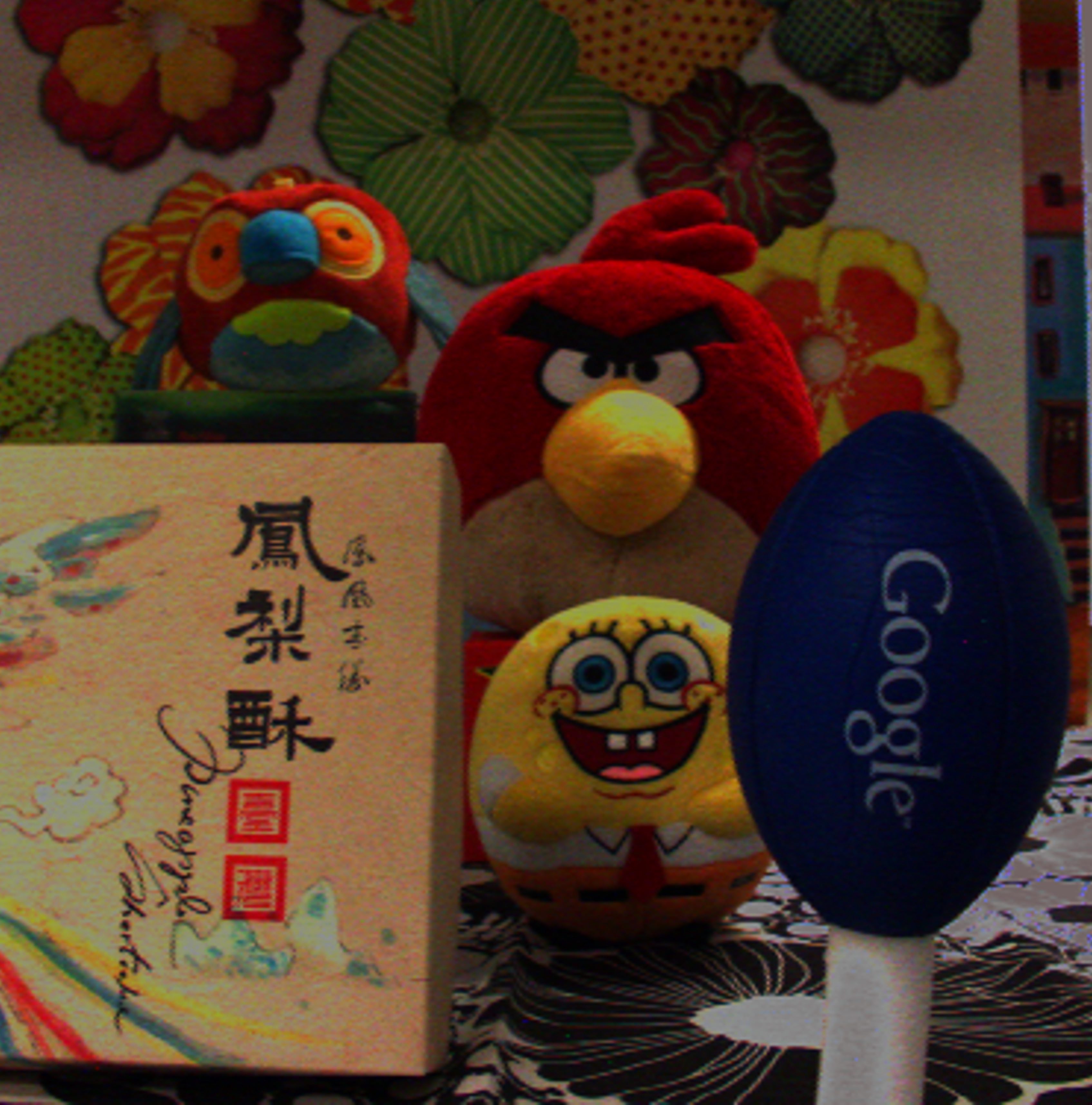}
\end{subfigure}
\begin{subfigure}{}
\includegraphics [trim={0 0 25 0},width=0.229\textwidth]{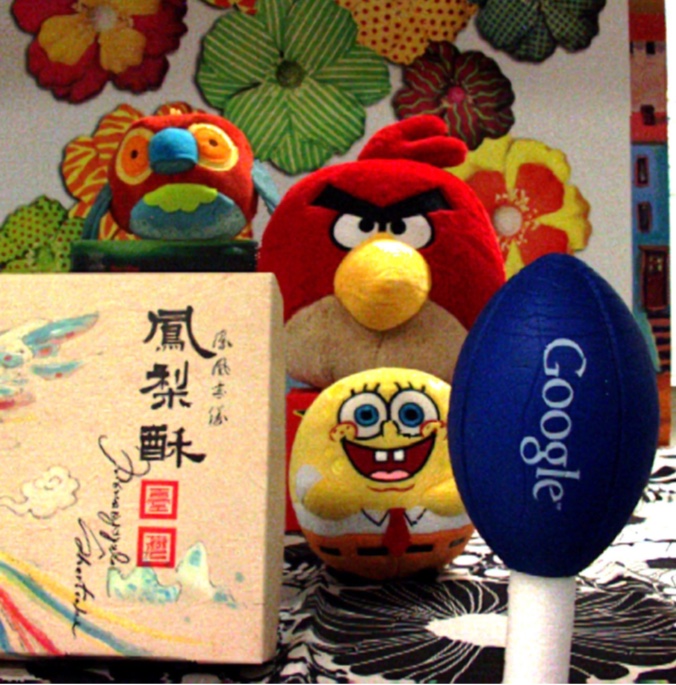}
\end{subfigure}\hspace{0.1in}
 \begin{subfigure}{}
\includegraphics [trim={0 0 35 0},width=0.229\textwidth]{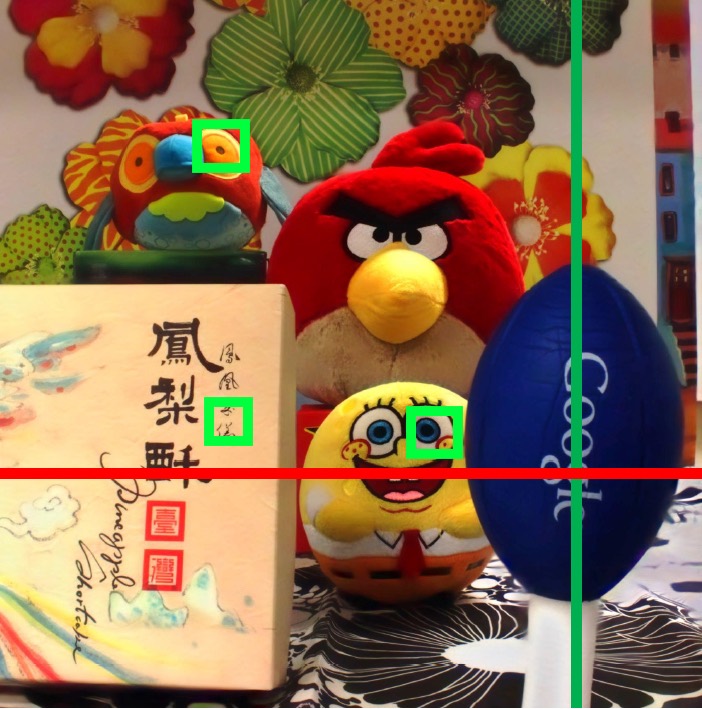}
\end{subfigure}
 \begin{subfigure}{}\hspace{0.1in}
\includegraphics [trim={0 0 0 0},width=0.24\textwidth]{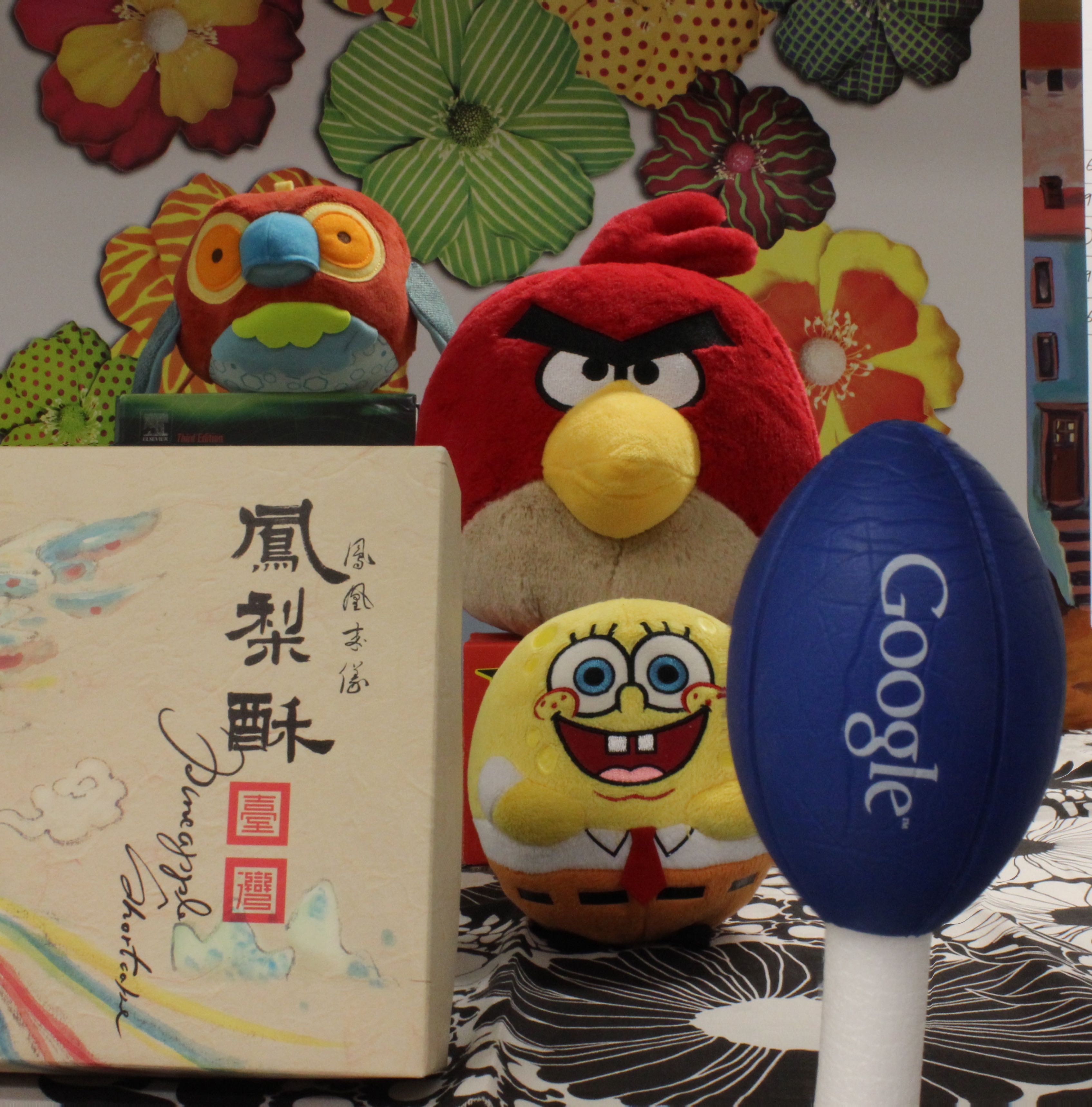}\\
\end{subfigure}
\label{fig:warped}
\hspace{0.3in}\textcolor{black}{ Dansereau et al. \cite{dansereau}}\hspace{0.6in}\textcolor{black}{ Cho et al. \cite{Cho}} \hspace{0.9in} \textcolor{black}{ Boominathan et al. \cite{Boominathan}}\hspace{0.5in} \textcolor{black}{Proposed method}\\
\caption{Comparison of various light field decoding and resolution enhancement methods.}
\label{fig:Comparison1}
\end{figure*}

\begin{figure*}
\centering
\hspace{-0.2in}
\begin{subfigure}{}
 \includegraphics[width=2cm,cfbox=red 1pt 1pt,trim={975 1270 2150 1899},clip,width=1.6in,height=1.6in,keepaspectratio]{Danserou}
 \end{subfigure}
\begin{subfigure}{}
\includegraphics[width=2cm,cfbox=blue 1pt 1pt,trim={0 0 0 0},clip,width=1.6in,height=1.6in,keepaspectratio]{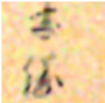}
\end{subfigure}
\begin{subfigure}{}
    \includegraphics[width=2cm,cfbox=purple 1pt 1pt,trim={0 0 0 0},clip,width=1.6in,height=1.6in,keepaspectratio]{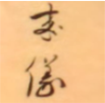}
\end{subfigure}
\begin{subfigure}{}
    \includegraphics[width=2cm,cfbox=black 1pt 1pt,trim={0 0 0 0},clip,width=1.6in,height=1.6in,keepaspectratio]{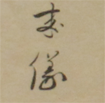}\\
\end{subfigure}
\hspace{-0.2in}
\begin{subfigure}{}
 \includegraphics[width=2cm,cfbox=red 1pt 1pt,trim={945 2590 2200 599},clip,width=1.6in,height=1.6in,keepaspectratio]{Danserou}
\end{subfigure}
\begin{subfigure}{}
\includegraphics[width=2cm,cfbox=blue 1pt 1pt,trim={0 0 0 0},clip,width=1.6in,height=1.6in,keepaspectratio]{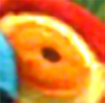}
  \end{subfigure}
\begin{subfigure}{}
\includegraphics[width=2cm,cfbox=purple 1pt 1pt,trim={0 0 0 0},clip,width=1.6in,height=1.6in,keepaspectratio]{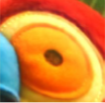}
  \end{subfigure}  
\begin{subfigure}{}
\includegraphics[width=2cm,cfbox=black 1pt 1pt,trim={15 0 2 5},clip,width=1.6in,height=1.51in]{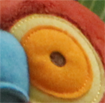}\\
  \end{subfigure}
\hspace{-0.12in}  
  \includegraphics[width=2cm,cfbox=red 1pt 1pt,trim={1985 1210 1150 1968},clip,width=1.6in,height=1.6in,keepaspectratio]{Danserou}\hspace{0.05in}
\begin{subfigure}{}
\includegraphics[width=2cm,cfbox=blue 1pt 1pt,trim={0 0 0 0},clip,width=1.6in,height=1.6in,keepaspectratio]{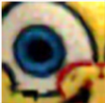}
  \end{subfigure}  
\begin{subfigure}{}
\includegraphics[width=2cm,cfbox=purple 1pt 1pt,trim={0 0 0 0},clip,width=1.6in,height=1.6in,keepaspectratio]{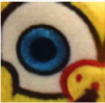}
\end{subfigure}  
\begin{subfigure}{}
\includegraphics[width=2cm,cfbox=black 1pt 1pt,trim={0 3 5 1},clip,width=1.6in,height=1.6in,keepaspectratio]{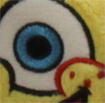}
  \end{subfigure}
\hspace{0.3in}\textcolor{black}{ Dansereau et al. \cite{dansereau}}\hspace{0.6in}\textcolor{black}{ Cho et al. \cite{Cho}} \hspace{0.9in} \textcolor{black}{ Boominathan et al. \cite{Boominathan}}\hspace{0.5in} \textcolor{black}{Proposed method}\\
  \caption{Zoomed-in regions from Figure \ref{fig:Comparison1}.}
\label{fig:Comparison1zoom}
\end{figure*}

\begin{figure}
\centering
\includegraphics[width=0.5\textwidth]{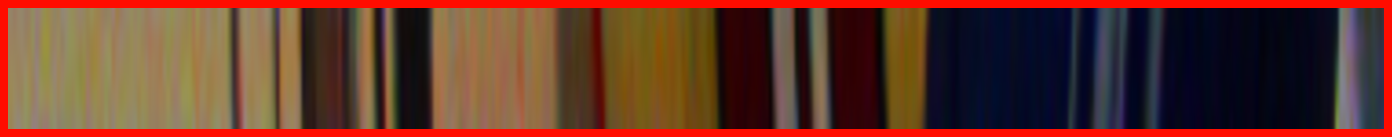}\vspace{-0.1cm}
\hspace{1.3in} \textcolor{black}{ Dansereau et al. \cite{dansereau} } 
\includegraphics[width=0.5\textwidth]{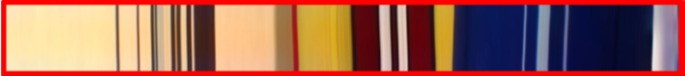}\vspace{-0.1cm}
 \hspace{1.3in} \textcolor{black}{ Boominathan et al. \cite{Boominathan} }  \\
\includegraphics[width=0.5\textwidth]{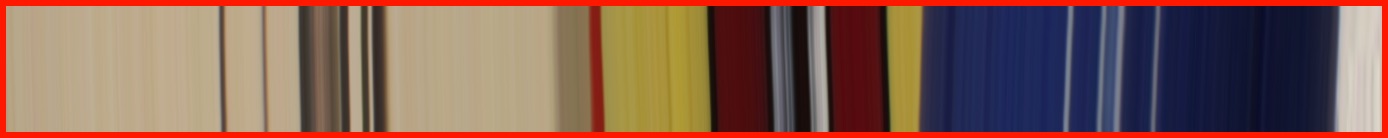}\vspace{-0.1cm}
\hspace{1.3in} \textcolor{black}{Proposed method}  \\
\includegraphics[width=0.501\textwidth]{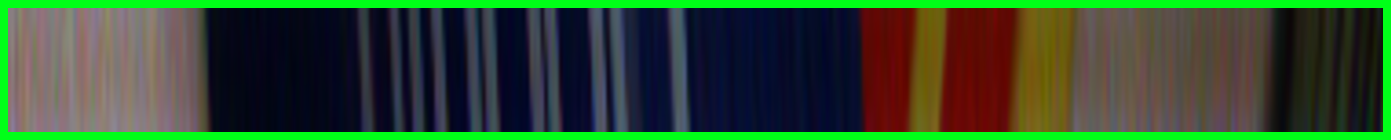}\vspace{-0.1cm}
\hspace{1.3in} \textcolor{black}{ Dansereau et al. \cite{dansereau} }  \\
\includegraphics[width=0.50\textwidth]{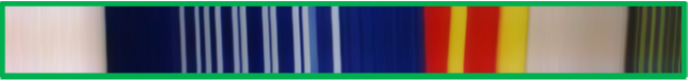}\vspace{-0.1cm}
\hspace{1.3in} \textcolor{black}{ Boominathan et al. \cite{Boominathan} } 
\includegraphics[width=0.5\textwidth]{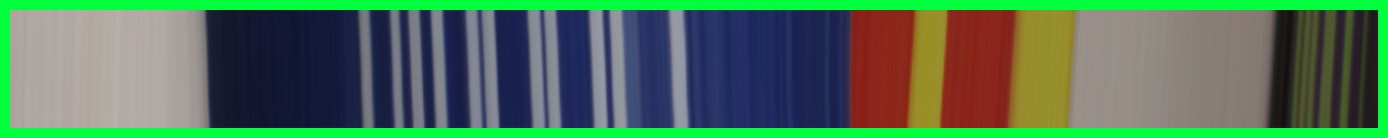}\vspace{-0.1cm}
\hspace{1.3in} \textcolor{black}{ Proposed method}  \\
\caption{Comparison of EPI images. The corresponding EPI lines are marked in Figure \ref{fig:Comparison1}.}
\label{fig:Comparison1EPI}
\end{figure}

In Figure \ref{fig:Comparison1}, we compare the proposed method with the hybrid imaging method of Boominathan et al. \cite{Boominathan}, the learning-based method of  Cho et al. \cite{Cho}, and the baseline decoder of Dansereau et al. \cite{dansereau}. Zoomed-in regions from these results are given in Figure \ref{fig:Comparison1zoom}. Comparing these results, we see that single-sensor methods cannot perform as well as hybrid methods. Among the hybrid methods, the proposed method produces sharper images than the method given in \cite{Boominathan}. In addition to producing sharper images, the proposed method has much less computational complexity than the one in \cite{Boominathan}, which takes about an hour on a similar PC configuration. Finally, we provide an epipolar-plane image (EPI) comparison in Figure \ref{fig:Comparison1EPI}. Again, the proposed method seems to be the best in preserving fine features.

\begin{figure*}
\centering
\includegraphics[trim={0 0 0 8cm},clip,width=2.0in]{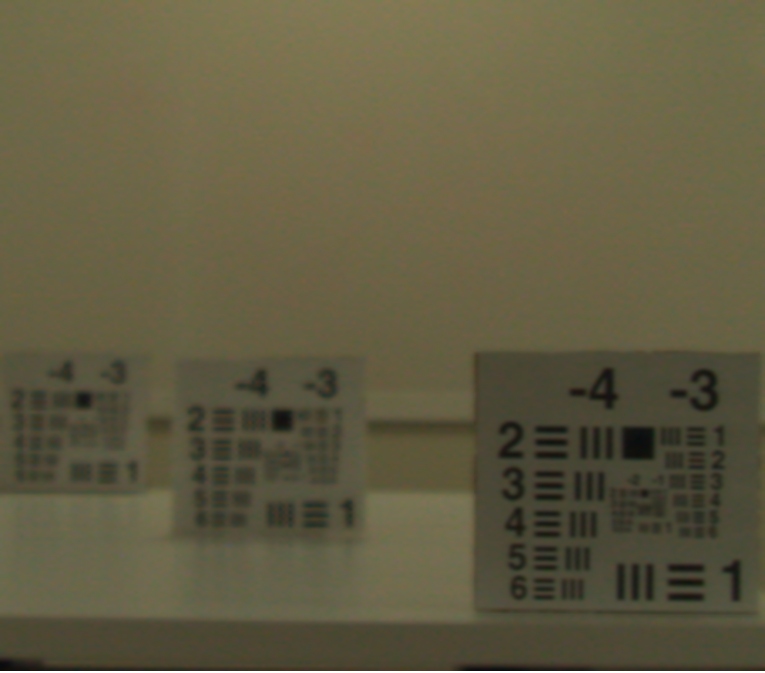}
\includegraphics[trim={0 0 0 8cm},clip,width=2.0in]{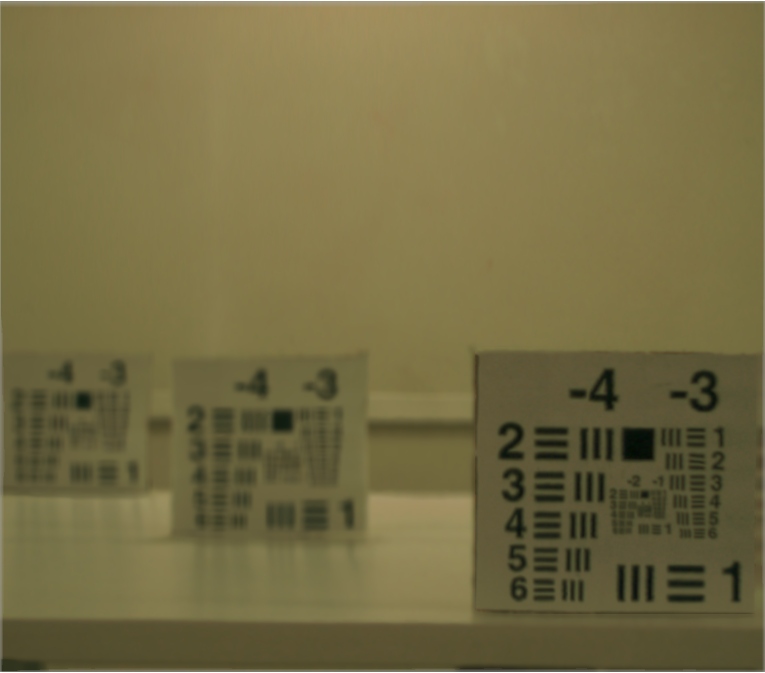}
\includegraphics[trim={0 0 0 8cm},clip,width=2.0in]{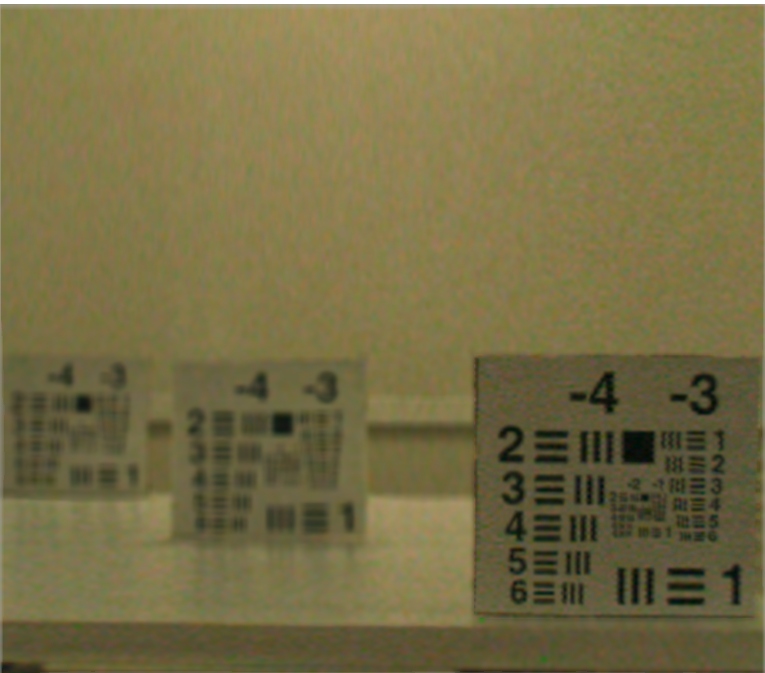}\\
\vspace{-3cm}
\textcolor{white}{[Close focus]} \hspace{1.3in} \textcolor{white}{[Close focus]} \hspace{1.3in} \textcolor{white}{[Close focus]}\\
\vspace{2.7cm}
\includegraphics[trim={0 0 0 8cm},clip,width=2.0in]{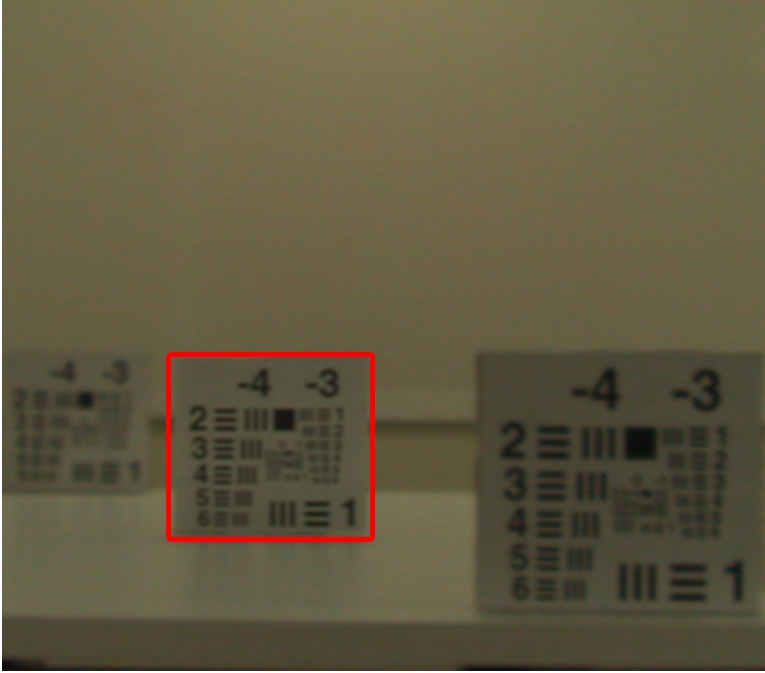}
\includegraphics[trim={0 0 0 8cm},clip,width=2.0in]{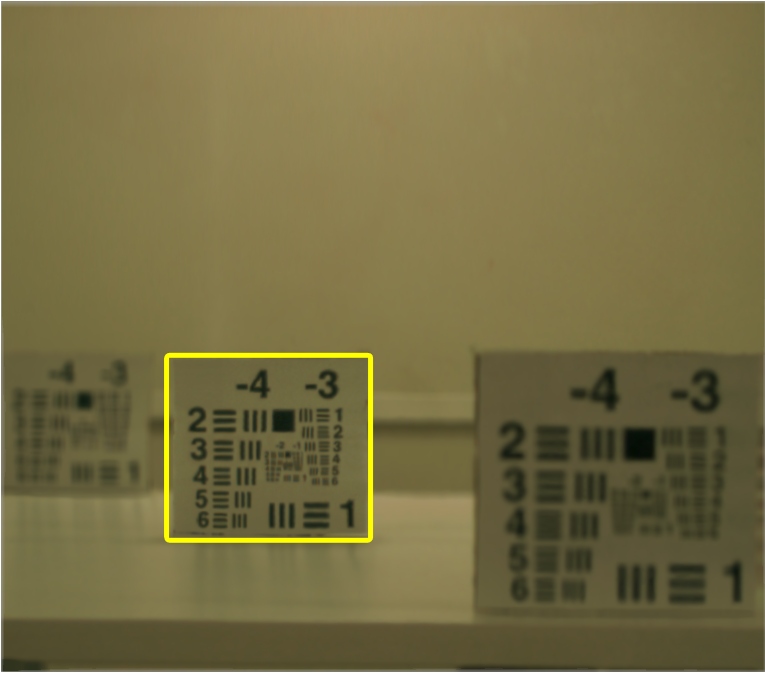}
\includegraphics[trim={0 0 0 8cm},clip,width=2.0in]{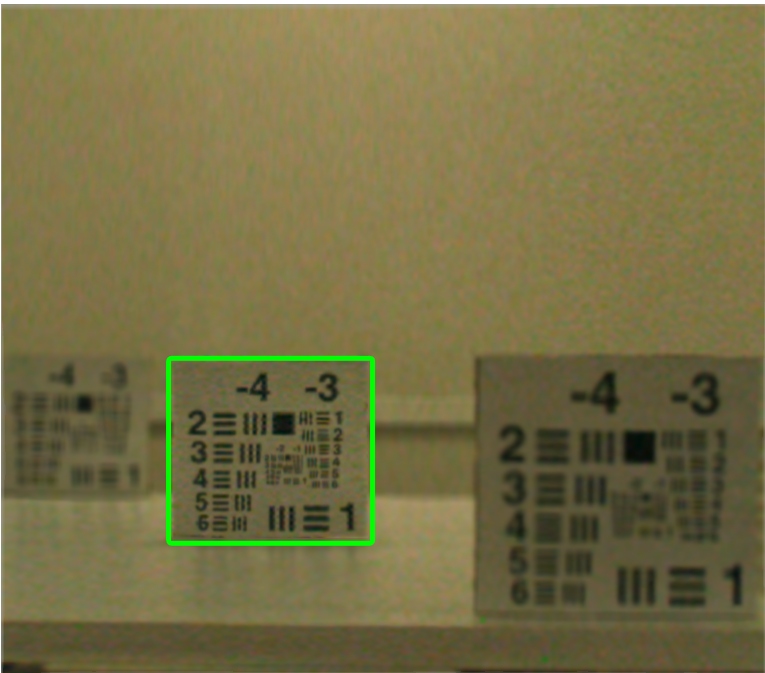}\\
\vspace{-3cm}
\textcolor{white}{[Mid focus]} \hspace{1.3in} \textcolor{white}{[Mid focus]} \hspace{1.3in} \textcolor{white}{[Mid focus]}\\
\vspace{2.7cm}
\includegraphics[trim={0 0 0 8cm},clip,width=2.0in]{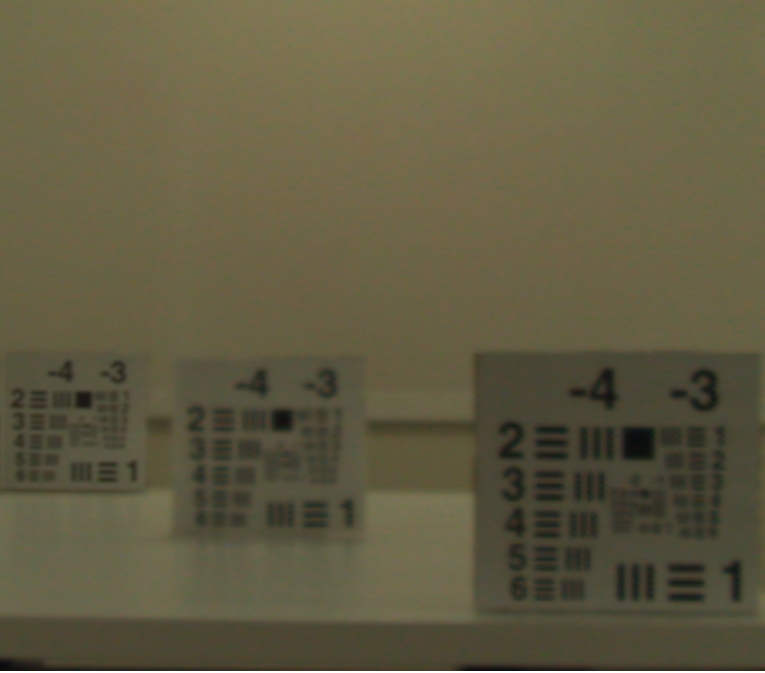}
\includegraphics[trim={0 0 0 8cm},clip,width=2.0in]{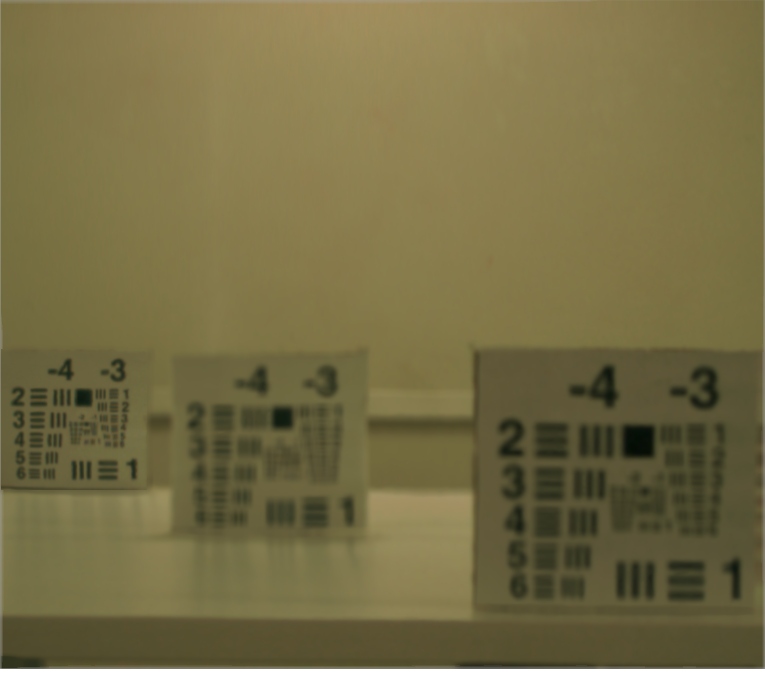}
\includegraphics[trim={0 0 0 8cm},clip,width=2.0in]{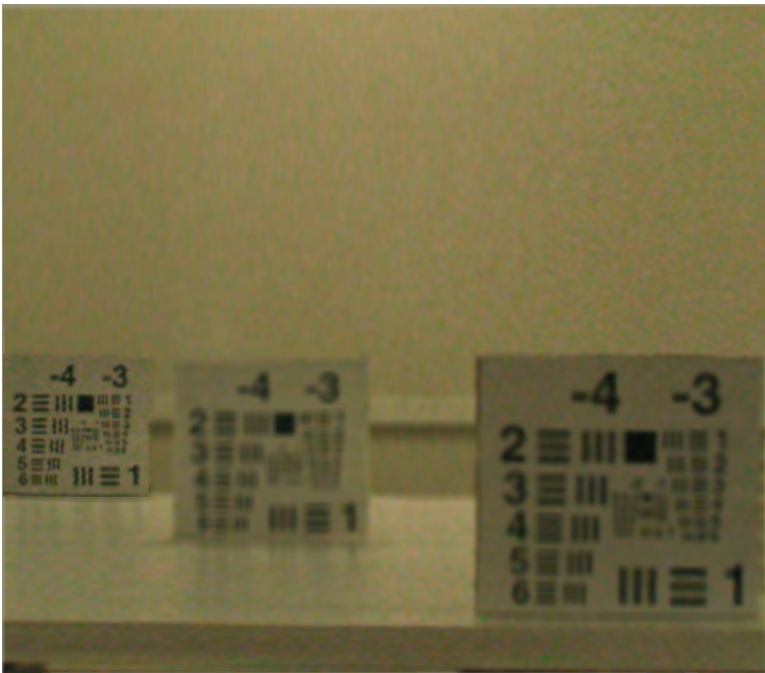}\\
\vspace{-3cm}
\textcolor{white}{[Far focus]} \hspace{1.3in} \textcolor{white}{[Far focus]} \hspace{1.3in} \textcolor{white}{[Far focus]}\\
\vspace{2.8cm}
\includegraphics[width=2.0in]{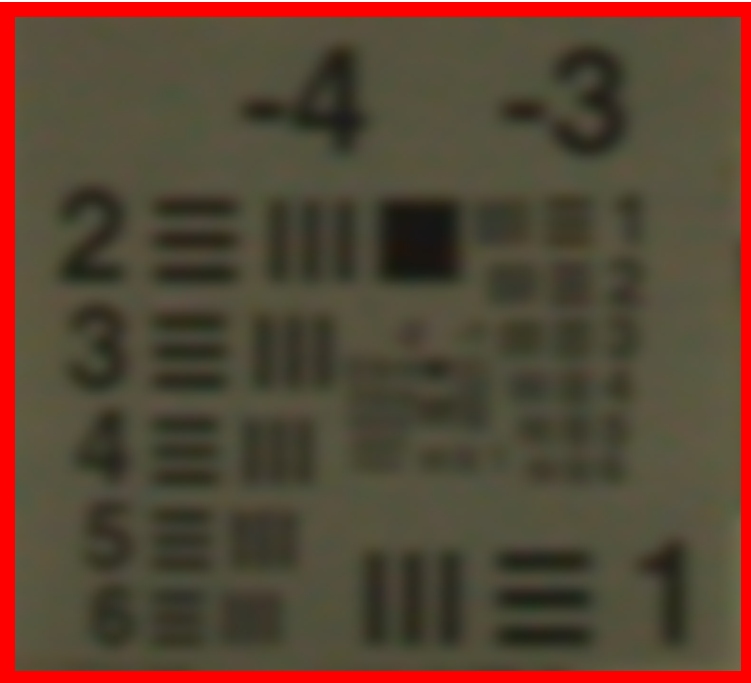}
\includegraphics[width=2.0in]{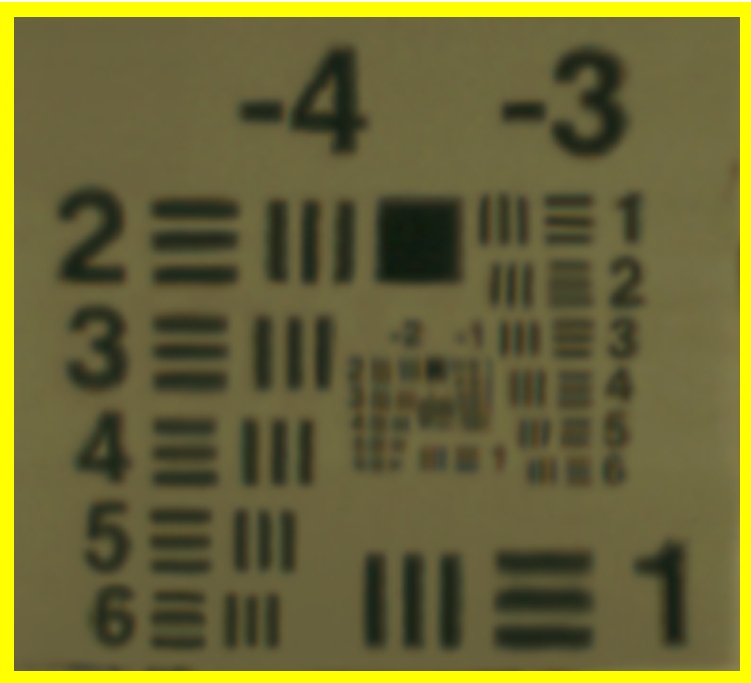}
\includegraphics[width=2.0in]{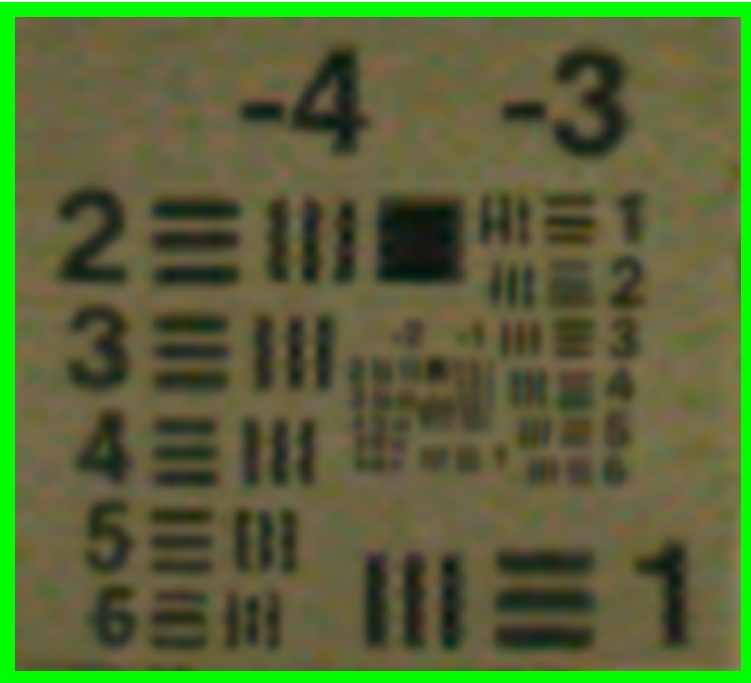}\\
(a) \hspace{1.8in} (b) \hspace{1.8in} (c)
\caption{Post-capture digital refocusing to close, middle and far depth using the shift-and-sum technique. (a) Single-sensor (Lytro) light field refocusing. (b) Resolution-enhanced (using alpha blending) light field refocusing. (c) Resolution-enhanced (using wavelet-based fusion) light field refocusing. The bottom row shows zoomed-in regions from middle-depth focusing.}
\label{fig:RF}
\end{figure*}

\begin{figure*}
\centering
\includegraphics[width=2.0in]{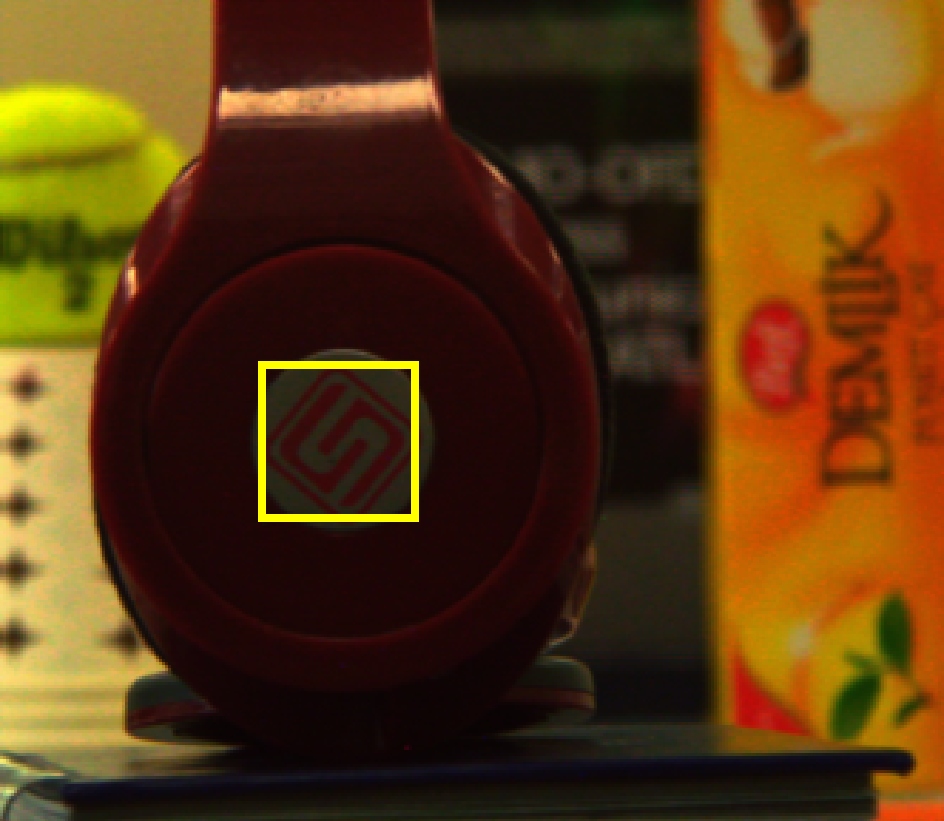}
\includegraphics[width=2.0in]{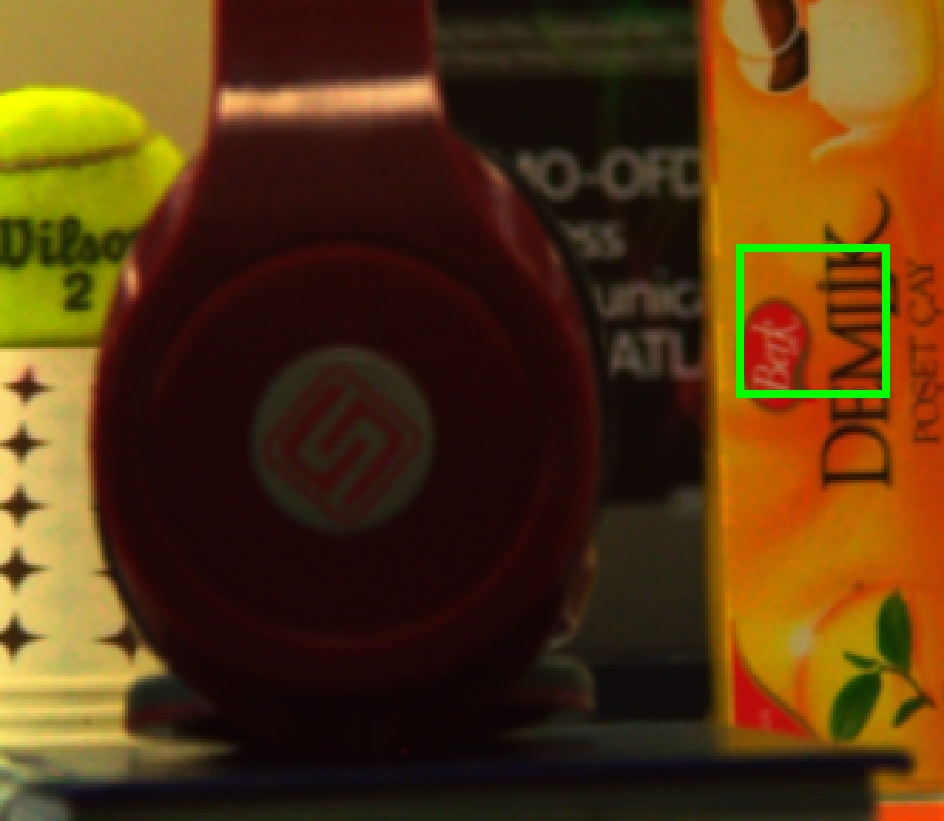}
\includegraphics[width=2.0in]{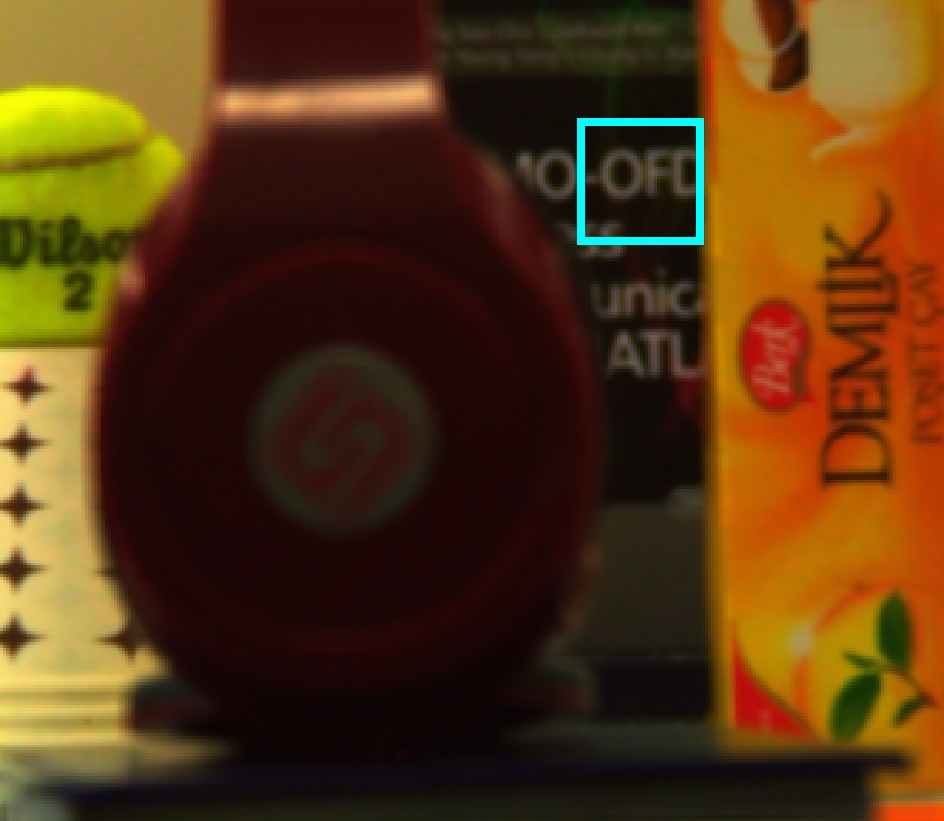}\\
\vspace{-4.45cm}
\textcolor{white}{[Lytro]} \hspace{1.6in} \textcolor{white}{[Lytro]} \hspace{1.6in} \textcolor{white}{[Lytro]}\\
\vspace{4.15cm}
\includegraphics[width=2.0in]{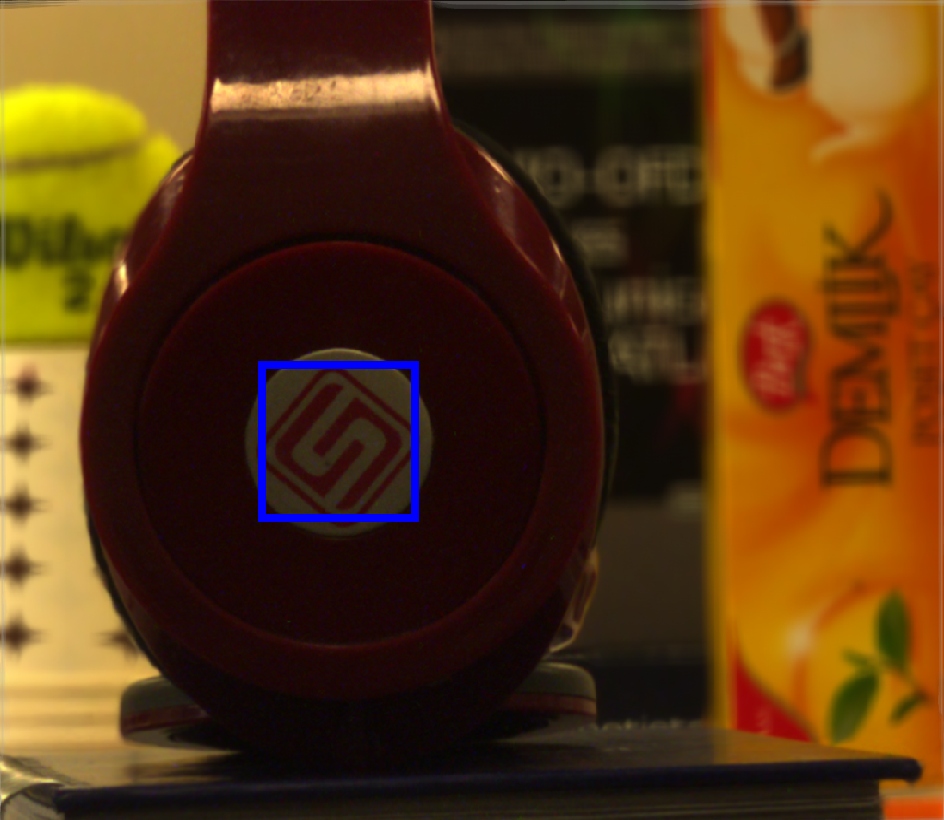}
\includegraphics[width=2.0in]{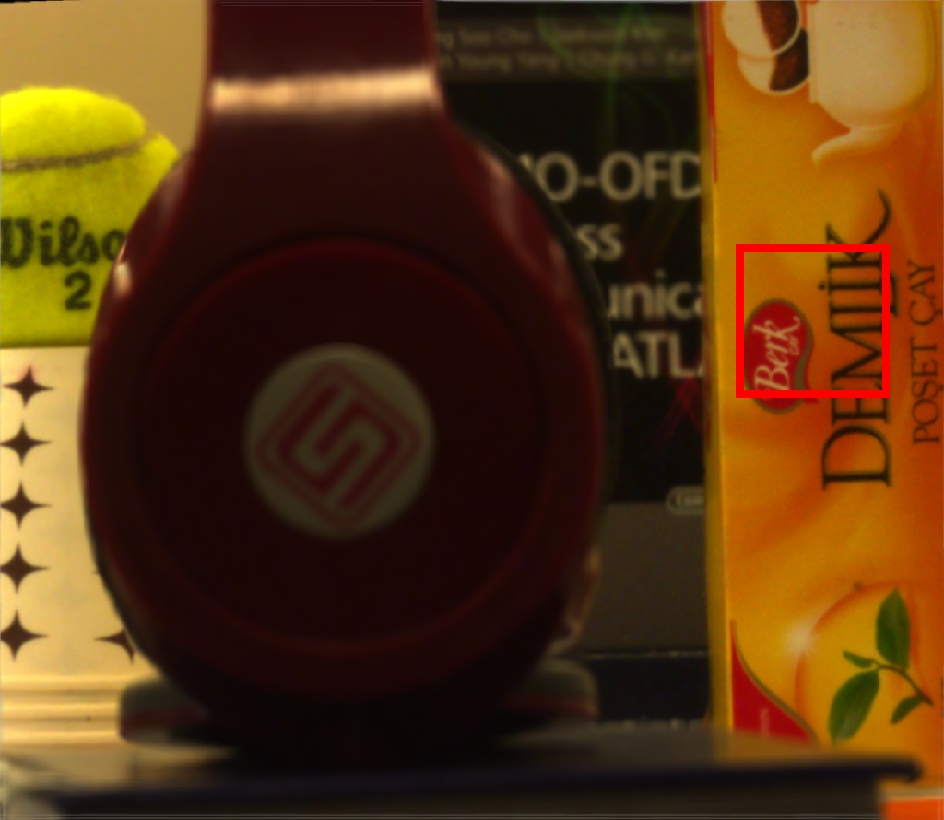}
\includegraphics[width=2.0in]{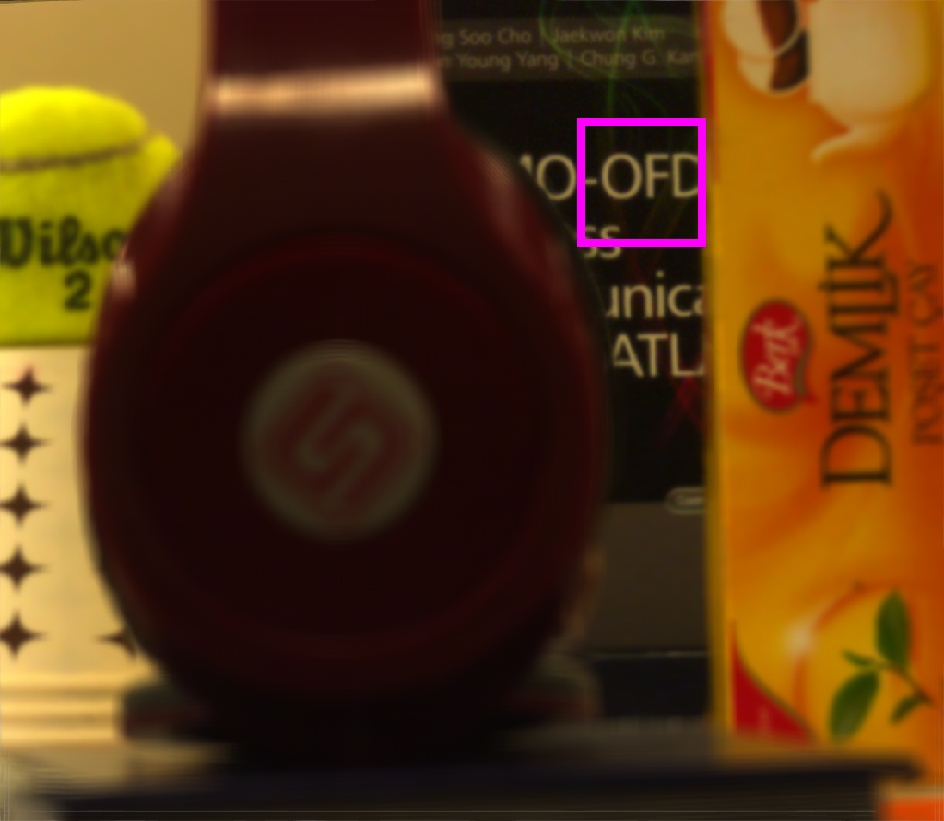}\\
\vspace{-4.45cm}
\textcolor{white}{[Resolution enhanced]} \hspace{0.7in} \textcolor{white}{[Resolution enhanced]} \hspace{0.7in} \textcolor{white}{[Resolution enhanced]}\\
\vspace{4.15cm}
\includegraphics[width=2.0in]{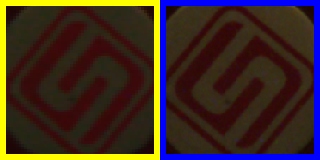}
\includegraphics[width=2.0in]{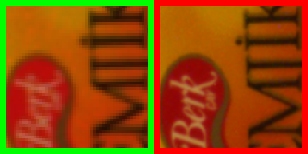}
\includegraphics[width=2.0in]{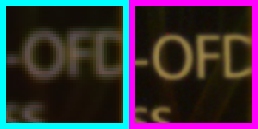}\\
(a) \hspace{1.8in} (b) \hspace{1.8in} (c)
\caption{Post-capture digital refocusing of single-sensor (Lytro) light field data and resolution-enhanced (using alpha blending) light field data using the shift-and-sum technique. (a) Close-depth focus. (b) Middle-depth focus. (c) Far-depth focus. The bottom row shows zoomed-in regions.}
\label{fig:RF2}
\end{figure*}

{\it Refocusing:} One of the key features of light field imaging is post-capture digital refocusing through a simple shift-and-sum procedure \cite{levoy}. In Figure \ref{fig:RF}, we show refocusing at different distances with Lytro light field images and the resolution-enhanced light field sub-aperture images. It can clearly be seen that we can obtain sharper refocusing compared to the original Lytro images. In Figure \ref{fig:RF2}, we provide refocusing examples from another data set captured by our imaging system. Again, the resolution-enhanced light field results in higher resolution refocused images compared to the Lytro light field.

\begin{figure*}
  \centering
  \includegraphics[width=1.8in]{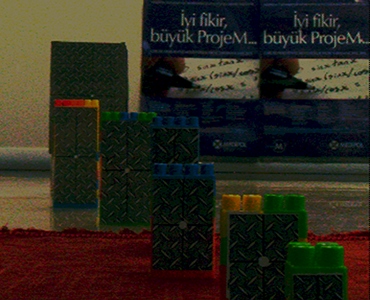}
  \includegraphics[width=1.8in]{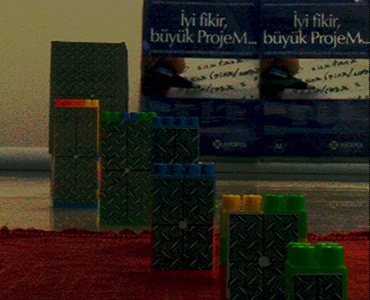}
  \includegraphics[width=1.8in]{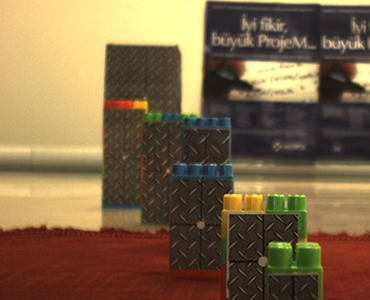}\\
  (a) \hspace{1.6in} (b) \hspace{1.6in} (c) \\

\includegraphics[width=1.55in, height=1.37in]{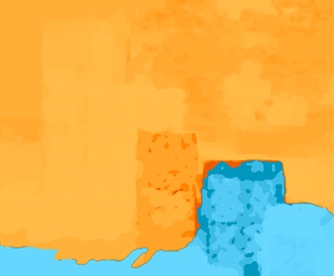}
\includegraphics[width=1.65in]{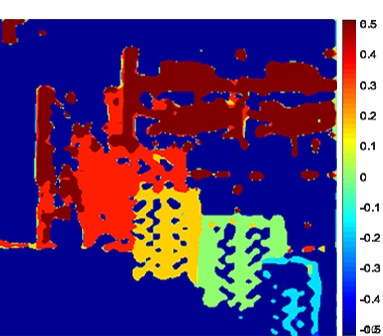}
\includegraphics[width=1.64in]{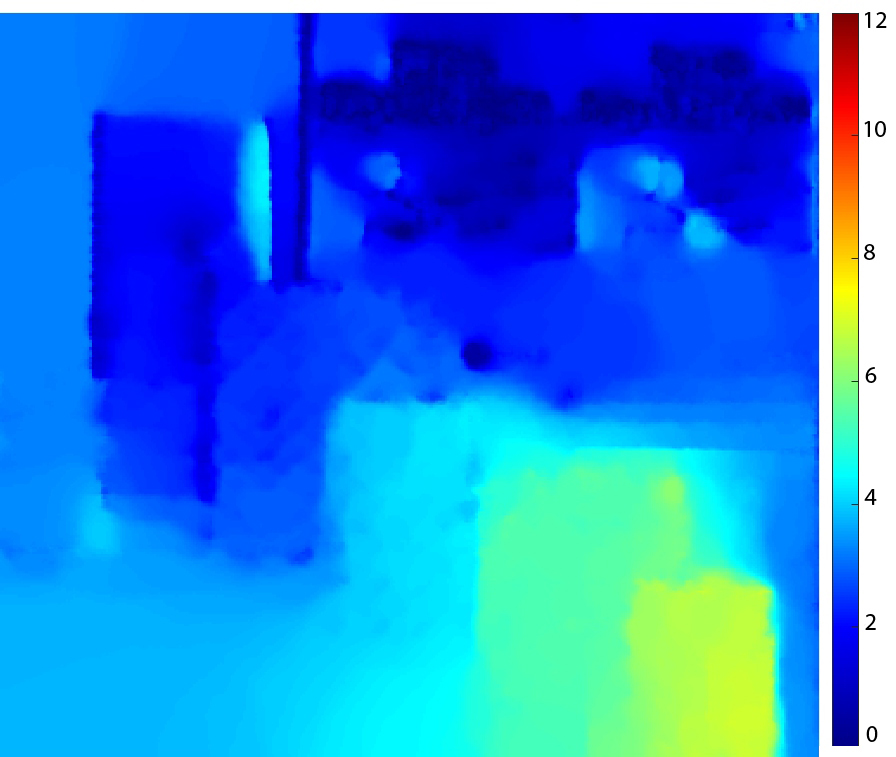}
\includegraphics[width=1.64in]{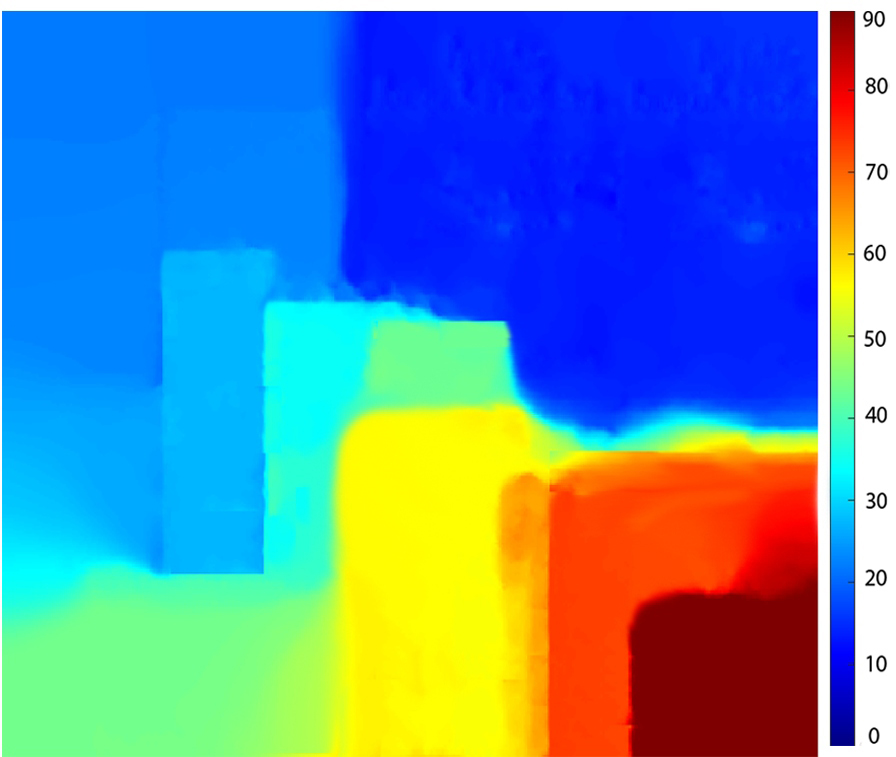}\\

 \hspace{-0.3in}(d)\hspace{1.5in}(e)\hspace{1.5in}(f)\hspace{1.5in} (g) \\
  \includegraphics[width=3 in]{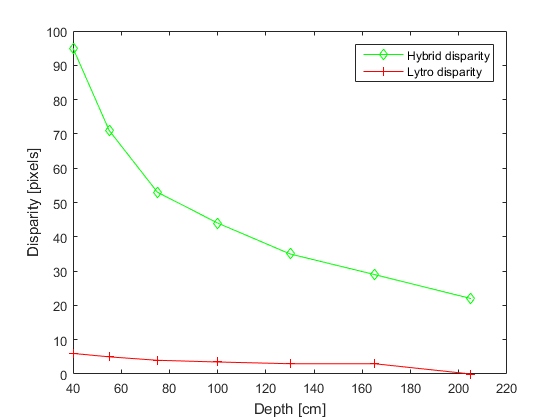}\\
  (h) \\
  \caption{Disparity map comparison. (a) Leftmost Lytro sub-aperture image. (b) Rightmost Lytro sub-aperture image. (c) Regular camera image (before photometric registration). (d) Disparity map of the Lytro software. (e) Disparity map of \cite{calderon2014depth}. (f) Disparity map between the leftmost and rightmost Lytro sub-aperture images. (g) Disparity map between the middle Lytro sub-aperture image and the regular camera image. (h) Disparities of the target object centers.}
   \label{fig:LRHRDISPARITY}
\end{figure*}

{\it Improving depth range and accuracy:} To demonstrate the increased depth range and improved depth estimation accuracy of our hybrid imaging system, we devised an experimental setup, where target objects (i.e, ``Lego blocks'') are placed in the scene starting from 40cm away from the imaging system. In Figure~\ref{fig:LRHRDISPARITY}, we show the leftmost and rightmost light field sub-aperture images as well as the regular camera image, in addition to the disparity maps estimated in different ways. Figure  \ref{fig:LRHRDISPARITY}(g) is the disparity map of the proposed hybrid system, computed between the regular camera image and the middle sub-aperture image using \cite{Liu}. Figure  \ref{fig:LRHRDISPARITY}(f) is the disparity map of light field camera, computed between the leftmost and rightmost sub-aperture images using \cite{Liu}. Figure  \ref{fig:LRHRDISPARITY}(e) is the disparity map estimated by \cite{calderon2014depth}, which uses all the sub-aperture images to estimate the disparity map and specifically designed for micro-lens array based light field cameras. And finally,  Figure  \ref{fig:LRHRDISPARITY}(d) is the disparity map produced by the Lytro manufacturer's proprietary software. Among these different approaches, we see that the proposed system produces the best disparity maps in distinguishing objects from different depths. Figure  \ref{fig:LRHRDISPARITY}(h),  the disparities of the target object positions are plotted. For the light field camera, the disparity difference from one depth to another becomes too small beyond 100cm, making it difficult to distinguish between different depths, and the disparities eventually become sub-pixel beyond 200cm. On the other hand, for the hybrid system, the disparities are large and distinguishable in the same range.

\begin{figure*}
\centering
\begin{subfigure}{}
 \begin{tikzpicture}
   \node[anchor=south west,inner sep=0] (image) at (0,0) {\includegraphics[width=0.19\textwidth]{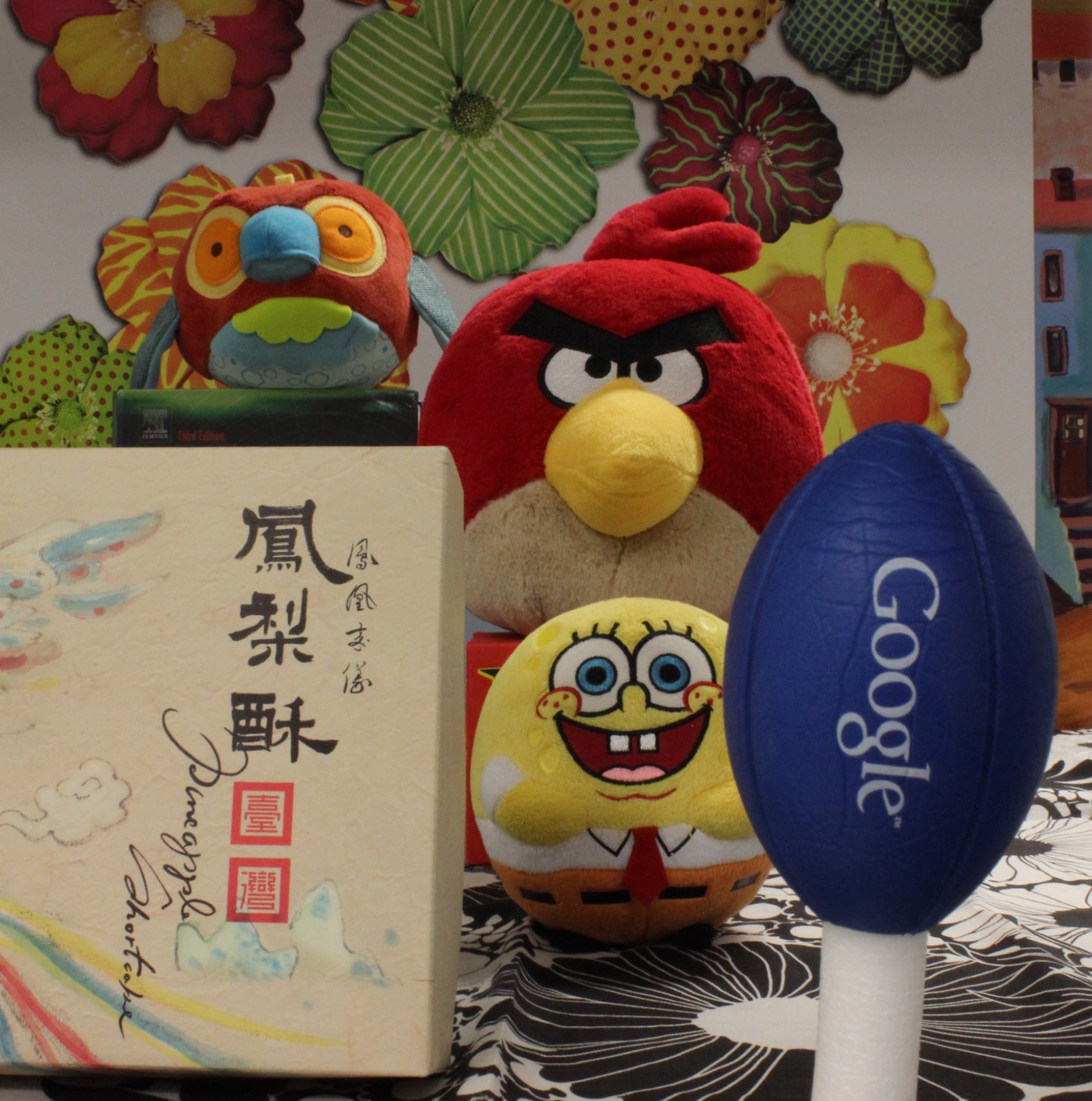}};
    \begin{scope}[x={(image.south east)},y={(image.north west)}]
        \draw[red,ultra thin,sharp corners] (0.8,0.58) rectangle (0.92,0.68);
    \end{scope}
     \begin{scope}[x={(image.south east)},y={(image.north west)}]
        \draw[yellow,ultra thin,sharp corners] (1,0.43) rectangle (0.85,0.56);
    \end{scope}
\end{tikzpicture}
  \end{subfigure}\hspace{-0.1in}
\begin{subfigure}{}
 \begin{tikzpicture}
   \node[anchor=south west,inner sep=0] (image) at (0,0) {\includegraphics[width=0.19\textwidth]{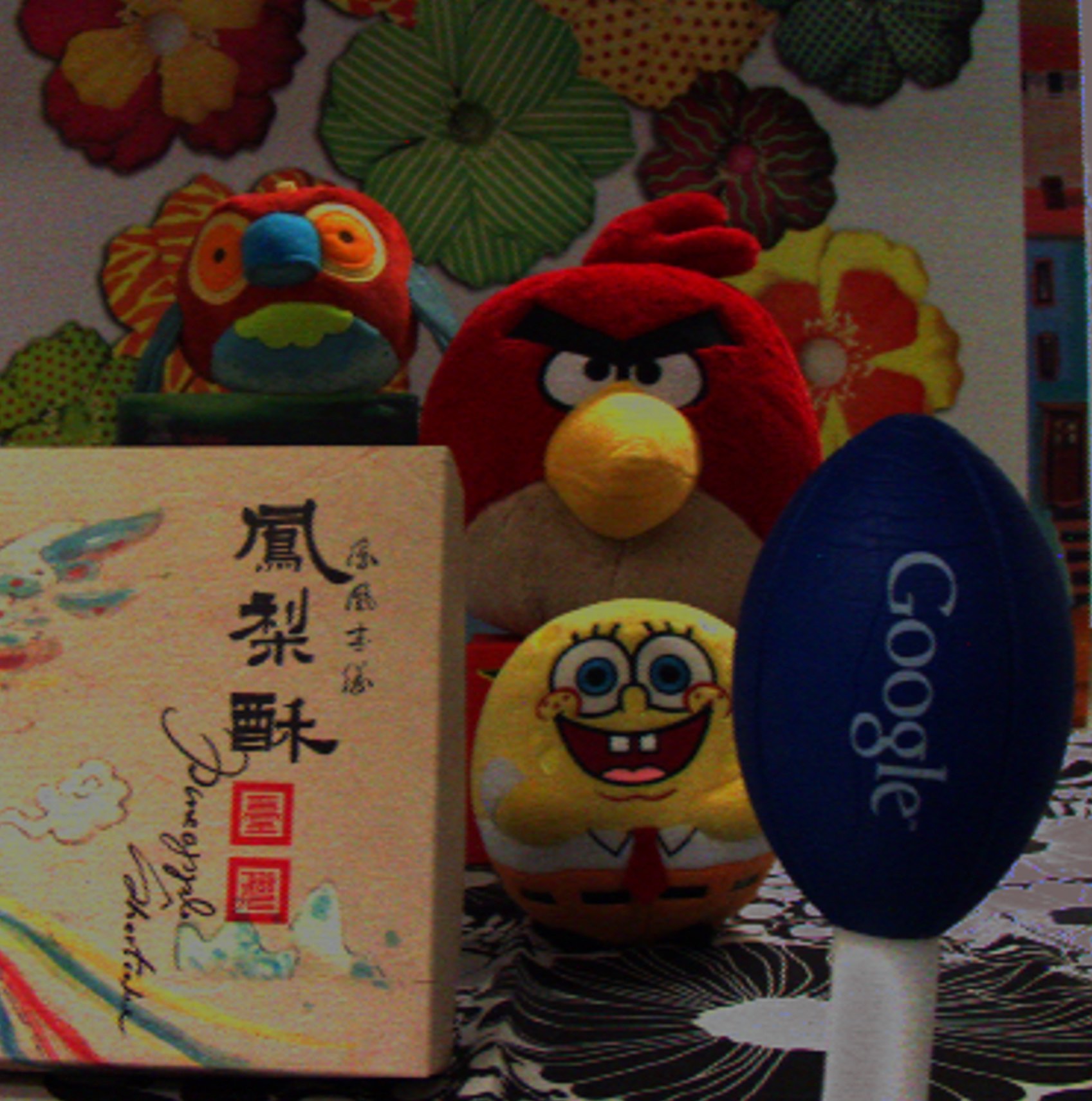}};
    \begin{scope}[x={(image.south east)},y={(image.north west)}]
        \draw[red,ultra thin,sharp corners] (0.8,0.58) rectangle (0.92,0.68);
    \end{scope}
     \begin{scope}[x={(image.south east)},y={(image.north west)}]
        \draw[yellow,ultra thin,sharp corners] (1,0.43) rectangle (0.85,0.56);
    \end{scope}
\end{tikzpicture}
  \end{subfigure}
\begin{subfigure}{}\hspace{-0.1in}
 \begin{tikzpicture}
   \node[anchor=south west,inner sep=0] (image) at (0,0) {\includegraphics[width=0.19\textwidth]{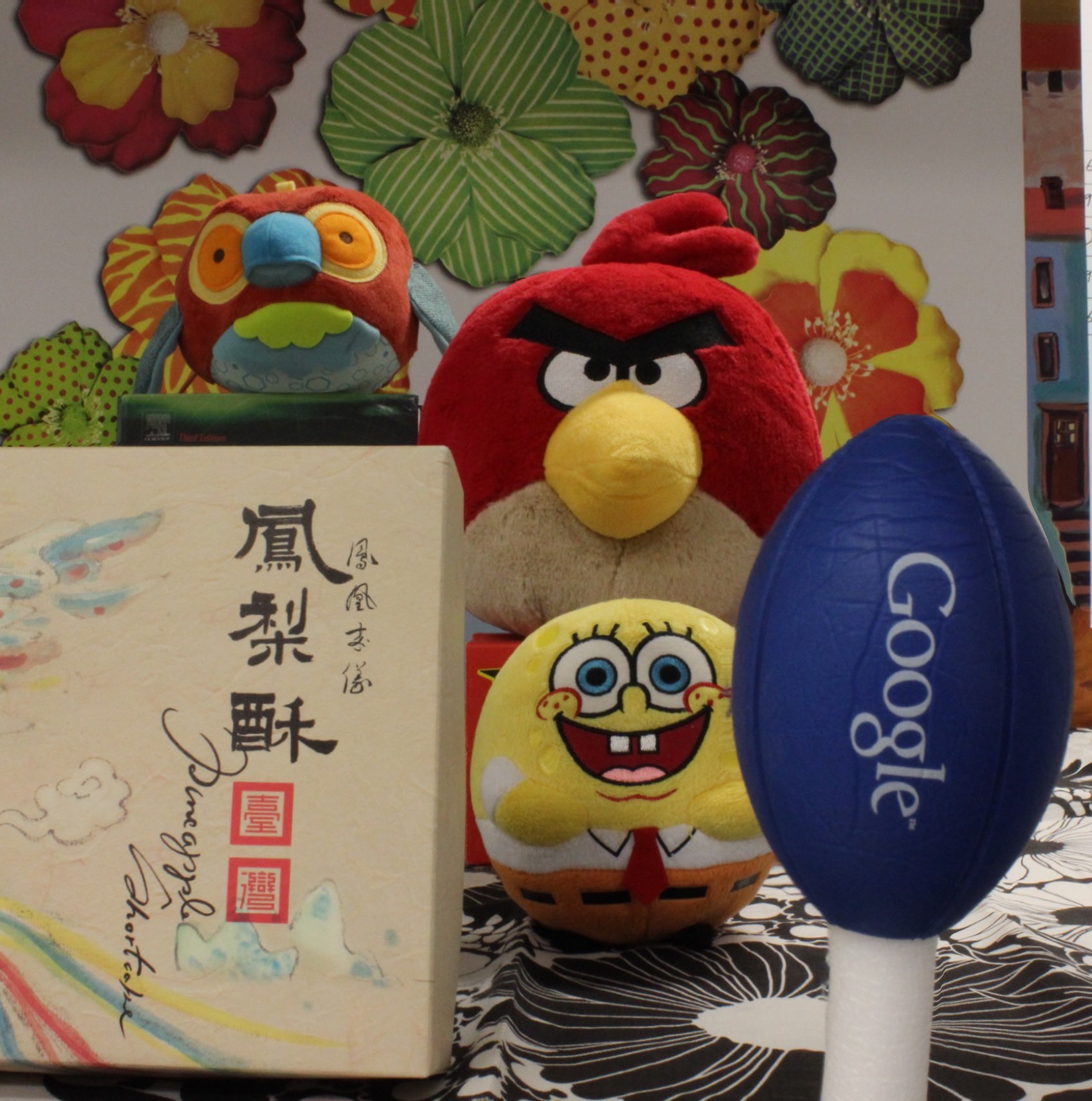}};
    \begin{scope}[x={(image.south east)},y={(image.north west)}]
        \draw[red,ultra thin,sharp corners] (0.8,0.58) rectangle (0.92,0.68);
    \end{scope}
     \begin{scope}[x={(image.south east)},y={(image.north west)}]
        \draw[yellow,ultra thin,sharp corners] (1,0.43) rectangle (0.85,0.56);
    \end{scope}
\end{tikzpicture}
  \end{subfigure}
\begin{subfigure}{}\hspace{-0.1in}
\begin{tikzpicture}
   \node[anchor=south west,inner sep=0] (image) at (0,0) {\includegraphics[width=0.19\textwidth]{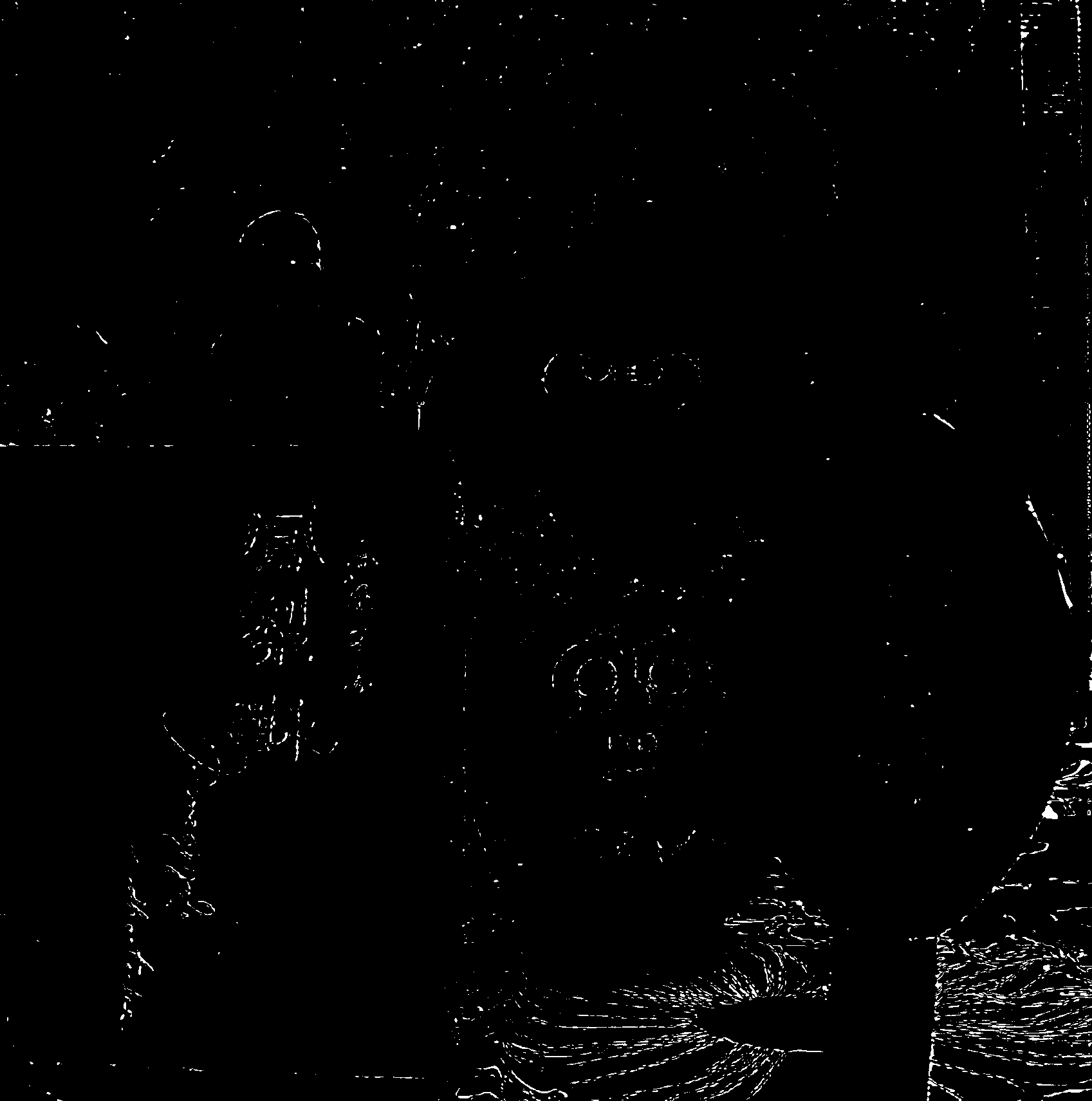}};
    \begin{scope}[x={(image.south east)},y={(image.north west)}]
        \draw[red,ultra thin,sharp corners] (0.8,0.58) rectangle (0.92,0.68);
    \end{scope}
     \begin{scope}[x={(image.south east)},y={(image.north west)}]
        \draw[yellow,ultra thin,sharp corners] (1,0.43) rectangle (0.85,0.56);
    \end{scope}
\end{tikzpicture}
  \end{subfigure} 
\begin{subfigure}{}\hspace{-0.1in}
\begin{tikzpicture}
   \node[anchor=south west,inner sep=0] (image) at (0,0) {\includegraphics[width=0.19\textwidth]{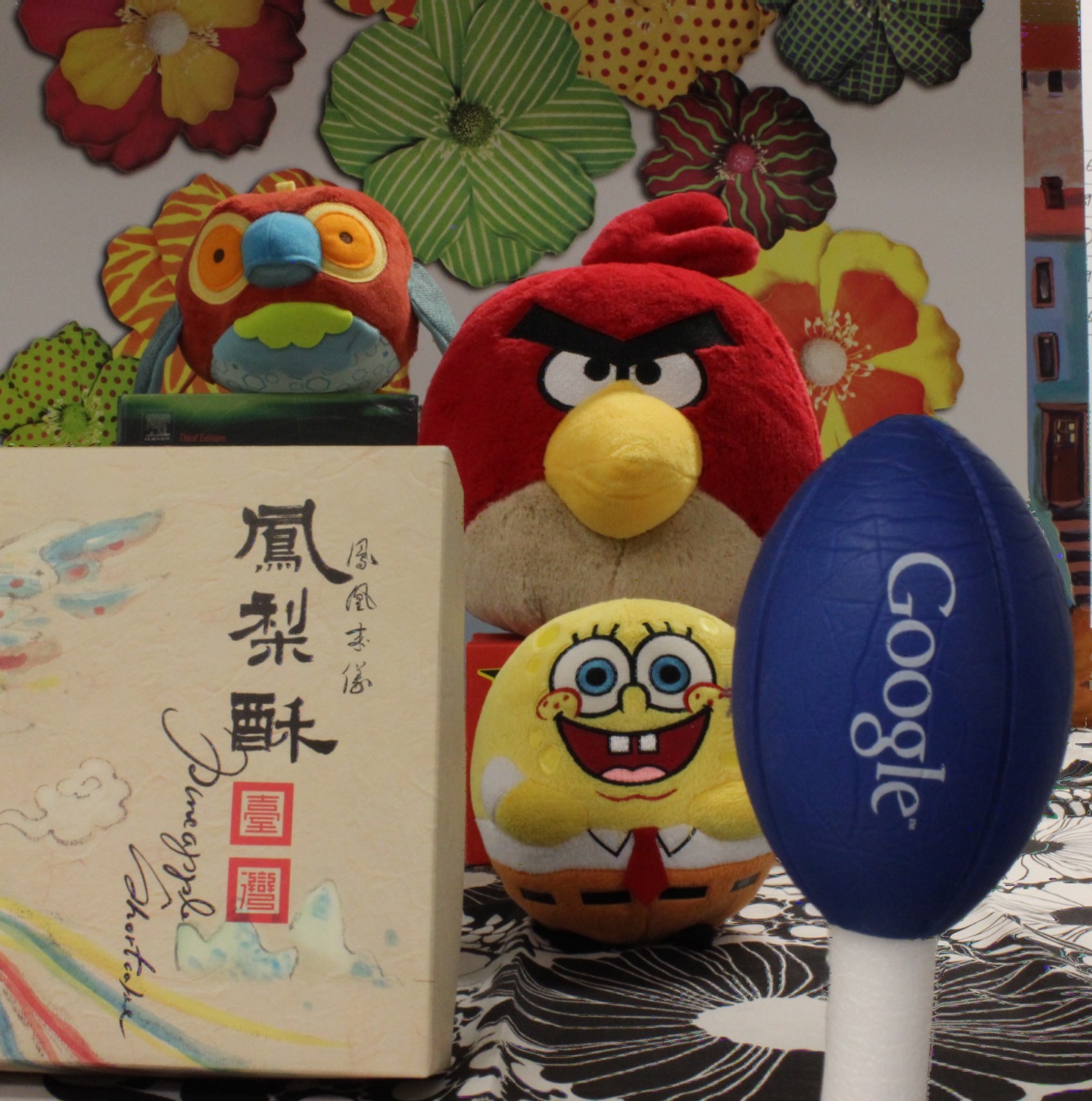}};
    \begin{scope}[x={(image.south east)},y={(image.north west)}]
        \draw[red,ultra thin,sharp corners] (0.8,0.58) rectangle (0.92,0.68);
    \end{scope}
     \begin{scope}[x={(image.south east)},y={(image.north west)}]
        \draw[yellow,ultra thin,sharp corners] (1,0.43) rectangle (0.85,0.56);
    \end{scope}
\end{tikzpicture}
\end{subfigure} 

\centering

\begin{subfigure}{}
 \includegraphics[width=2cm,cfbox=red 1pt 1pt,trim={1299 930 90 540},clip,width=1.2in]{Hrr}
\end{subfigure}
\begin{subfigure}{}
    \includegraphics[width=2cm,cfbox=red 1pt 1pt,trim={1299 930 90 540},clip,width=1.2in]{Lres}
\end{subfigure}
\begin{subfigure}{}
    \includegraphics[width=2cm,cfbox=red 1pt 1pt,trim={1299 930 90 540},clip,width=1.2in]{Warped}
\end{subfigure}
\begin{subfigure}{}
\includegraphics[width=2cm,cfbox=red 1pt 1pt,trim={1299 930 90 540},clip,width=1.2in]{FM}
  \end{subfigure}
\begin{subfigure}{}
\includegraphics[width=2cm,cfbox=red 1pt 1pt,trim={1299 930 90 540},clip,width=1.2in]{fused}\\
  \end{subfigure}
\begin{subfigure}{}
\includegraphics[width=2cm,cfbox=yellow 1pt 1pt,trim={1420 780 0 740},clip,width=1.2in]{Hrr}
  \end{subfigure}  
\begin{subfigure}{}
\includegraphics[width=2cm,cfbox=yellow 1pt 1pt,trim={1420 780 0 740},clip,width=1.2in]{Lres}
  \end{subfigure}  
  \begin{subfigure}{}
\includegraphics[width=2cm,cfbox=yellow 1pt 1pt,trim={1420 780 0 740},clip,width=1.2in]{Warped}
  \end{subfigure}  
\begin{subfigure}{}
\includegraphics[width=2cm,cfbox=yellow 1pt 1pt,trim={1420 780 0 740},clip,width=1.2in]{FM}
  \end{subfigure} 
\begin{subfigure}{}
\includegraphics[width=2cm,cfbox=yellow 1pt 1pt,trim={1420 780 0 740},clip,width=1.2in]{fused}\\
  \end{subfigure} 
 
  \textcolor{black} {\hspace{0.0in} (a) \hspace{1.1in} (b) \hspace{1.1in} (c) \hspace{1.1in} (d) \hspace{1.1in} (e)} 
  
\caption{Occlusion handling. (a) High-resolution regular camera image. (b) Lytro light field sub-aperture image. (c) Resolution enhanced light field sub-aperture image. (d) Occlusion mask. (e) Occluded regions are filled from Lytro sub-aperture image. }
\label{fig:Holefilling}
\end{figure*}

{\it Addressing occlusion:} The proposed method, which registers high-resolution image and light field sub-aperture images using optical flow estimation, can handle occluded regions to a large extent. This is demonstrated in Figure \ref{fig:Holefilling}. In Figure \ref{fig:Holefilling}(a), the high-resolution regular camera image is shown. A low-resolution Lytro sub-aperture image is given in Figure \ref{fig:Holefilling}(b). In Figure \ref{fig:Holefilling}(c), the resolution-enhanced version of the sub-aperture image is provided. As seen in the zoomed-in regions, the proposed method can handle the occlusion well in most parts as the optical flow based registration aims to minimize the brightness difference. However, there are still some regions, where the difference between the resolution enhanced and low-resolution input is large. One method to detect these occluded regions is to compare the absolute difference against a threshold; when the difference in any color channel is larger than a pre-determined threshold, we can mark the corresponding pixel as occluded. Figure \ref{fig:Holefilling}(d) shows the occlusion mask when the threshold is set to 0.175 after some trial and error, with the pixel values in the range 0 to 1. The occluded regions can then be filled with the pixel values from the original light field image, as shown in Figure \ref{fig:Holefilling}(e). In this approach, the value of the threshold is critical. Choosing the threshold too small may cause missing the occlusion regions; on the other hand, a large threshold value may lead to transferring noisy and low-resolution data into the final image. 

\section{Conclusions}

In this paper, we presented a hybrid imaging system that includes a light field camera and a regular camera. The system, while keeping the capabilities of light field imaging, improves both spatial resolution and depth estimation range/accuracy due to increased baseline. Because the fixed stereo system allows pre-calibration and by utilizing the fact that light field sub-aperture images are captured on a regular grid, the registration of low-resolution light field sub-aperture images and high-resolution regular camera images is simplified. With proper image registration, even a simple image fusion, such as alpha blending, produces good results. The method compares favorably against several methods in the literature, including two single-image light field decoders and a hybrid-camera light field resolution enhancer.   



\end{document}